\documentclass[twoside,leqno,twocolumn]{article}

\usepackage[letterpaper]{geometry}

\usepackage{siamproceedings}

\usepackage[T1]{fontenc}
\usepackage{amsfonts}
\usepackage{graphicx}
\usepackage{epstopdf}
\usepackage{enumitem}
\usepackage{algorithmic}
\ifpdf
  \DeclareGraphicsExtensions{.eps,.pdf,.png,.jpg}
\else
  \DeclareGraphicsExtensions{.eps}
\fi

\newsiamremark{remark}{Remark}
\newsiamremark{hypothesis}{Hypothesis}
\crefname{hypothesis}{Hypothesis}{Hypotheses}
\newsiamthm{claim}{Claim}

\usepackage{printlen}

\usepackage{caption}
\usepackage{booktabs}
\usepackage{tabularx}

\usepackage{xcolor}

\usepackage{amsopn}

\begin{document}

\newcommand\relatedversion{}
\renewcommand\relatedversion{\thanks{The full version of the paper can be accessed at \protect\url{https://arxiv.org/abs/0000.00000}}} % Replace URL with link to full paper or comment out this line

% \title{A Unified and Scalable Framework for Collision-based Proximities in Tree Ensembles through Leaf Incidence Maps}
% \title{A Unified and Scalable Framework for Tree Ensemble Proximities via Leaf Incidence Maps}
% \title{Forest Proximities as Sparse Leaf-Incidence Kernels
% }
\title{Revisiting Forest Proximities via Sparse Leaf-Incidence Kernels}

    \author{Adrien Aumon\thanks{Département de mathématiques et de statistique, Université de Montréal | Mila -- Quebec AI Institute, Montreal, Quebec, Canada.}
    \and Guy Wolf\footnotemark[1]    
    \and Kevin R. Moon\thanks{Department of Mathematics and Statistics,
Utah State University, Logan, Utah, USA.}
  \and Jake S. Rhodes\thanks{Department of Statistics,
Brigham Young University, Provo, Utah, USA
(\email{rhodes@stat.byu.edu}).}}

    % \author{Anonymous Authors}

\date{}

\maketitle

% Copyright Statement
% When submitting your final paper to a SIAM proceedings, it is requested that you include
% the appropriate copyright in the footer of the paper.  The copyright added should be
% consistent with the copyright selected on the copyright form submitted with the paper.
% Please note that "20XX" should be changed to the year of the meeting.

% Default Copyright Statement
\fancyfoot[R]{\scriptsize{Copyright \textcopyright\ 2026 by SIAM\\
Unauthorized reproduction of this article is prohibited}}

% Depending on which copyright you agree to when you sign the copyright form, the copyright
% can be changed to one of the following after commenting out the default copyright statement
% above.

%\fancyfoot[R]{\scriptsize{Copyright \textcopyright\ 20XX\\
%Copyright for this paper is retained by authors}}

%\fancyfoot[R]{\scriptsize{Copyright \textcopyright\ 20XX\\
%Copyright retained by principal author's organization}}

%\pagenumbering{arabic}
%\setcounter{page}{1}%Leave this line commented out.

\begin{abstract}%   <- trailing '%' for backward compatibility of .sty file

Decision forests induce supervised similarities through the partition structure of their trees. Yet forest proximity computation is still often treated as a quadratic operation in the number of samples, which limits scalability and restricts broader use in kernel and representation-learning pipelines. We introduce a unified view of leaf-collision forest proximities through a class of Separable Weighted Leaf-Collision (SWLC) kernels, showing that most existing proximities differ only in their weighting scheme while sharing a common sparse leaf-incidence structure. This yields an explicit leaf-space representation that clarifies their kernel interpretation and leads to an exact finite-sample sparse factorization of the proximity matrix, avoiding an explicit all-pairs comparison and reducing computation to sparse linear algebra over leaf collisions. We implement this framework in a memory-efficient Python library and show, both theoretically and empirically, that exact kernel computation scales near-linearly in time and memory under standard forest regimes. Benchmarks verify the predicted scaling behavior in practice across datasets, proximity definitions, and forest settings, and show that the resulting sparse leaf-space representation can also be used directly for fast task-aware embedding.
Our code is available at \url{https://github.com/JakeSRhodesLab/ForestGeom}.

\end{abstract}

\section{Introduction.}

% Beyond their widespread use for classification and regression, Random Forests (RFs;~\cite{breiman_random_2001, breiman_manual_2003}) and other decision-forest methods such as Extremely Randomized Trees (ExtraTrees;~\cite{geurts_extremely_2006}) and Gradient Boosted Trees (GBTs;~\cite{friedman_greedy_2001}) have attracted growing interest for their ability to induce pairwise similarity measures from the partitioning structure of their trees [TO STRENGTHEN THIS CLAIM, ADD SEVERAL CITATIONS HERE]. This notion of \emph{forest proximity} was originally introduced by Breiman for RFs as the proportion of trees in which two observations fall into the same leaf~\cite{breiman_manual_2003}. Because splits are optimized for a supervised objective, these proximities yield a supervised similarity measure that implicitly reflects variable importance relative to auxiliary information.

Beyond their widespread use for classification and regression, Random Forests (RFs;~\cite{breiman_random_2001, breiman_manual_2003}) and other decision-forest methods such as Extremely Randomized Trees (ETs;~\cite{geurts_extremely_2006}) and Gradient Boosted Trees (GBTs;~\cite{friedman_greedy_2001}) also induce pairwise similarity measures through the partitioning structure of their trees. This notion of \emph{forest proximity} was originally introduced by Breiman for RFs as the proportion of trees in which two observations fall into the same leaf~\cite{breiman_manual_2003}. Because splits are optimized for a supervised objective, these proximities yield a supervised similarity measure that implicitly reflects variable importance relative to auxiliary information.

Forest proximities have since attracted growing interest as a flexible supervised similarity for both data representation and model understanding across a wide range of tasks. They have been used for outlier detection, imputation, and general model exploration~\cite{rhodes_geometry-_2023}, and to induce task-aware geometries for supervised dimensionality reduction (DR), visualization, and alignment, including guided biological data exploration through 2D embeddings of multiple sclerosis subtype trajectories~\cite{rhodesGainingBiologicalInsights2026}, RF-based autoencoding for out-of-sample embedding and reconstruction~\cite{aumon2025random, vu2025autoencoding}, supervised manifold alignment~\cite{rhodes_random_2024}, and RF biplots for variable interpretation in high-dimensional omics data~\cite{blanchet_constructing_2020}. They have also been used directly for prediction and decision making in prototype identification~\cite{tan_tree_2020}, tabular classification through RF-induced graph neural networks~\cite{farokhi_advancing_2024}, time-series classification via proximity forests~\cite{shaw_forest_2026}, patient-specific prediction in ICU data through local modeling~\cite{lee_patient-specific_2017}, supervised similarity learning in finance~\cite{jeyapaulraj_supervised_2022}, and matching treated and control subjects in causal inference~\cite{zhao_propensity_2016}. Because proximities expose both instance-level relationships and feature-level structure, they are also useful for interpretability and explanation, complementing attribution methods such as SHAP~\cite{lundberg_unified_2017} and LIME~\cite{ribeiro_why_2016}, with applications including visual feature contribution analysis~\cite{whitmore_explicating_2018}, prediction-faithful explanations of RF and GBT models~\cite{rhodes_geometry-_2023, geertsema_instance-based_2023, rosaler_enhanced_2024}, and the characterization of copy-number-variation structure among pea aphid host races in evolutionary genomics~\cite{duvaux_dynamics_2015}.

Forest proximities are often used as implicit supervised kernel matrices in spectral and kernel-based methods. While a dot-product or feature-map view has been noted in some cases~\cite{vu2025autoencoding, whitmore_explicating_2018, olson_making_2018}, there is still no general framework that explains when forest proximities admit explicit leaf-space representations, or how to use this structure in a systematic way for both methodology and scalability. Moreover, forest proximities remain hard to use at scale because many implementations require quadratic time or memory. Rhodes et al.~\cite{rhodesGainingBiologicalInsights2026} suggested that, for a fixed training point, one could restrict the computation to samples that share a leaf with it across trees, rather than compare it to all training samples, but they left this idea for future work.

We develop a general framework for forest proximities. We introduce a family of leaf-based similarities called \emph{Separable Weighted Leaf-Collision} (SWLC) proximities, and show that it includes most existing collision-based forest proximities. We also show that each member of this family has an explicit leaf-space representation, leading to an exact finite-sample sparse factorization of the proximity matrix in which computation is restricted to samples that collide in leaves. Under standard forest regimes, this yields near-linear time and space complexity in the number of samples. We support these results with a memory-efficient Python implementation and a broad empirical study across datasets, forest configurations, and kernel variants. The experiments confirm the expected runtime and memory trends in practice, and show that exact proximity computation remains feasible on datasets with hundreds of thousands of samples on standard CPU hardware. We also show that the same factorization enables the direct use of sparse leaf representations in downstream spectral methods, extending the value of forest proximities beyond scalable matrix construction alone.

\section{Background.}

\subsection{Overview of leaf-collision proximities.}\label{subsec:overview}

Leaf-collision proximities are a family of similarity measures built from sample co-occurrence in terminal nodes across a tree ensemble. The guiding principle is simple: two points should be  similar if they repeatedly fall into the same leaves across trees. At each node, the learning algorithm selects a split according to an optimization criterion such as entropy, Gini impurity, or mean squared error. Thus the routing path of a sample reflects a sequence of decisions based on features that are informative for the prediction task. This means that two similar samples followed the same routing path through a sequence of feature-based splits in multiple trees, and, therefore, remained indistinguishable under the task-adaptive partition learned by the ensemble, despite potential differences in uninformative features.  Leaf collisions thus capture a notion of proximity that emphasizes task-relevant structure rather than label-free geometry.

The original RF proximity introduced by Breiman~\cite{breiman_random_2001,breiman_manual_2003} measures similarity as the fraction of trees in which two samples fall into the same terminal leaf. This simple leaf-collision principle remains widely used and competitive in practice~\cite{farokhi_advancing_2024}. Kernel variants such as KeRF~\cite{scornet2016random} refine this definition by downweighting collisions in large leaves, yielding symmetric kernels with favorable properties for downstream methods such as diffusion~\cite{vu2025autoencoding}. Other extensions incorporate training-specific information. In particular, OOB proximities~\cite{liaw2002classification,Hastie2009} restrict collisions to trees where both samples are out-of-bag (OOB), reducing biases induced by bootstrap sampling and improving reliability in practice.

Building on these ideas, RF-GAP~\cite{rhodes_geometry-_2023} combines OOB querying with leaf-mass normalization and in-bag multiplicities to better align proximities with the predictive geometry of the forest, achieving strong empirical performance across tasks while introducing asymmetry that can be symmetrized if needed~\cite{rhodes_geometry-_2023,rhodesGainingBiologicalInsights2026,aumon2025random}. Other variants instead reweight collisions using external criteria, such as instance hardness in RFProxIH~\cite{cao_novel_2021}. Extensions to boosted trees have also been proposed, either through empirical per-tree weighting~\cite{tan_tree_2020} or by deriving instance similarities directly from the additive regression model~\cite{geertsema_instance-based_2023}, although the resulting weights are generally signed and are therefore better interpreted as instance-wise influence scores than as conventional nonnegative proximities. Refer to Appendix~\ref{app:overview_extended} for a more comprehensive literature review on forest proximities.

Each proposed proximity variant is motivated by a distinct goal, such as mitigating overfitting or faithful model explanations. However, these definitions have never been established on a common formal foundation: they are typically presented as stand-alone algorithmic recipes compared through downstream performance. In particular, it is often left implicit that collision-based proximities induce \emph{kernels} that can be written as inner products in a high-dimensional leaf space. Moreover, different proximity measures largely amount to different choices of leaf-incidence feature maps (and associated weights) prior to taking a dot product. This perspective not only unifies their definitions but also has important implications for how these proximities can be computed and scaled. In particular, aside from recent work on GPU acceleration or heuristic block-sparse approximations~\cite{kuchar_rfx_2025}, there has been limited effort to analyze the faithful computational cost of these inner-product constructions. As a result, collision-based proximities are often treated as inherently quadratic, $\mathcal{O}(N^2)$, in the number of samples, an assumption that has long discouraged their use at scale. This paper challenges that view by introducing a unifying algebraic framework that makes the underlying feature maps explicit, and by providing the corresponding complexity analysis needed to scale these measures to millions of samples.

\subsection{General assumptions and notations.}

Let $\mathcal{D}=(X,Y)=\{(x_i,y_i)\}_{i=1}^N \subset \mathcal{X}\times\mathcal{Y}$ be a finite labeled training set. Rather than working on a specific forest training algorithm, we suppose the existence of an \emph{ensemble context} $(\mathcal{T},\theta)$ generated from $\mathcal{D}$. The first component, $\mathcal{T}=\{\mathcal{T}_t\}_{t=1}^T$, denotes the \emph{ensemble topology}—the collection of decision tree structures that determine how samples are routed to leaves. The second component, $\theta$, is a fixed \emph{context object} providing auxiliary information related to the trained ensemble's geometry, such as training hyperparameters, tree-wise descriptors, leaf-level statistics, or algorithm-dependent quantities derived during or after training.

\paragraph{Leaf indexing.}
For each tree $t\in\{1,\ldots,T\}$, let $L_t$ denote its number of leaves, and let
$
L := \sum_{t=1}^T L_t
$ be the total number of leaves across the ensemble. The leaf index sets $\{\mathcal L_t\}_{t=1}^T$ form a partition of the global index set $\mathcal L = \{1,\ldots,L\}$, i.e.,
$\bigsqcup_{t=1}^T
\mathcal L_t = \mathcal{L},
$
with $|\mathcal L_t| = L_t$. The ensemble topology $\mathcal T$ then induces, for each tree $t$, a leaf index map
$
\ell_t:\mathcal X \to \mathcal L_t,
$
where $\ell_t(x)$ is the unique global index of the leaf reached by $x$ in tree $t$. Evaluating $\ell_t(x)$ requires traversing a single root-to-leaf path in tree $t$, and therefore costs $\mathcal O(h_t)$, where $h_t$ denotes the depth of tree $t$.

\paragraph{Weight assignment $\boldsymbol{q}$.}
We define a weighting map
\[
\boldsymbol{q} : \mathcal{X} \to \mathbb{R}^T_{\geq 0}, \qquad \boldsymbol{q}(x)=(q_{1}(x),\ldots,q_{T}(x)),
\]
where $q_t(x)$ is a nonnegative scalar weight associated with sample $x$ in tree $t$. 
In many proximity definitions, weights are determined by the fixed context and the leaf reached by $x$ (more precisely, one could write $q_t(x \mid \theta,\ell_t(x))$). For readability, and to allow for arbitrary per-sample, per-tree weighting schemes, we simply write $q_t(x)$. Intuitively, $\boldsymbol{q}(x)$ encodes the contribution (or importance) of $x$ across the ensemble. In typical constructions, evaluating $q_t(x)$ reduces to retrieving leaf-level statistics from $\theta$ in constant time. Our definition is deliberately more general, and allows $q_t$ to involve nontrivial computations when needed. For example, in RFProxIH~\cite{cao_novel_2021}, $q_t$ depends on an instance-hardness score that may require preprocessing or local neighborhood queries. Thus, defining $\boldsymbol{q}$ directly as a map on $\mathcal{X}$ keeps the weights locally tied to the ensemble partition while supporting flexible, sample-dependent assignments.

\section{Methods.}

We now introduce a unifying framework for forest proximities. By explicitly decoupling the topological collision from the contextual importance of each tree, we define a broad class of kernels that encompasses the majority of existing forest-based similarity measures.

\subsection{SWLC proximities.}

\begin{definition}[Separable Weighted Leaf-Collision (SWLC) Proximity]
\label{def:swlcp}
Given an ensemble context $(\mathcal{T}, \theta)$ and two weight assignments $\boldsymbol{q}$ and $\boldsymbol{w}$, a \emph{separable weighted leaf-collision} (SWLC) proximity is a function $P_{\boldsymbol{q},\boldsymbol{w}} : \mathcal{X} \times \mathcal{X} \to \mathbb{R}_{\geq 0}$ of the form:
\begin{equation}
\label{eq:swlcp}
P_{\boldsymbol{q},\boldsymbol{w}}(x,x') = \sum_{t=1}^T q_t(x) \, w_t(x') \, \mathbb{I}(\ell_t(x) = \ell_t(x')).
\end{equation}
\end{definition}

The term \emph{separable} highlights that the weights assigned to the first argument $x$ and the second argument $x'$ are determined independently sample-wise within each tree. In the special case where $\boldsymbol q=\boldsymbol w$, the proximity becomes symmetric and, as shown later, admits a Gram representation in a sparse leaf-induced feature space.

\begin{proposition}[Generality of SWLC proximities]\label{prop:swlc}
All leaf-collision based proximities introduced in Section~\ref{subsec:overview} are symmetric or asymmetric SWLC proximities, except OOB.
\end{proposition}

\begin{proof}
The proof is obtained by identifying the specific ensemble context and weight assignments $\boldsymbol{q}$ and $\boldsymbol{w}$ required to recover each measure. For detailed derivations of these weights from their canonical definitions, we refer the reader to Appendix~\ref{app:swlcp_proofs}.
\end{proof}

\begin{remark}
Although OOB proximities are not separable and therefore do not belong to the SWLC family, they remain symmetric leaf-collision proximities and retain the near-linear computational complexity discussed in Section~\ref{subsec:complexity}. See Appendix~\ref{subsec:oob} for details.
\end{remark}

\subsection{Sparse forest kernels.}

While the summation in Eq.~\eqref{eq:swlcp} provides a conceptual definition shared across most proximity definitions in the existing literature, its naive evaluation for all pairs $(x_i, x_j)$ is computationally prohibitive. To reveal the underlying geometry and enable scalable computation, we translate the SWLC proximity into a high-dimensional dot product. As discussed in Section~\ref{subsec:overview}, while the literature contains various sophisticated proximities, their formulations are often presented as algorithmic procedures where the underlying inner-product structure remains implicit. We make this structure explicit by representing each sample as a sparse vector in the global leaf space.

\begin{definition}[Weighted leaf-incidence map]\label{def:weighted_leaf_representations}
Given a weight assignment $\boldsymbol q$, the weighted leaf-incidence map
$\boldsymbol{\phi}_{\boldsymbol q}:\mathcal X\to\mathbb R^{L}_{\geq 0}$ is defined by
\begin{equation*}
\boldsymbol{\phi}_{\boldsymbol q}(x)
=
\sum_{t=1}^T q_t(x)\,\mathbf e_{\ell_t(x)},
\end{equation*}
where $\mathbf e_j\in\mathbb R^{L}$ denotes the $j$-th canonical basis vector.
\end{definition}

\begin{lemma}[$T$-Sparsity]\label{cor:l0_weighted}
Let $\|\cdot \|_0$ denote the number of nonzero elements ($L_0$ ``norm''). For any $x\in\mathcal X$, the representation $\boldsymbol{\phi}_{\boldsymbol q}(x)$ satisfies
\begin{equation}
\|\boldsymbol{\phi}_{\boldsymbol q}(x)\|_0 = \|\boldsymbol q(x)\|_0 \le T.
\end{equation}
\end{lemma}

\begin{proof}
By Definition~\ref{def:weighted_leaf_representations}, we have
\begin{align*}
\|\boldsymbol{\phi}_{\boldsymbol q}(x)\|_0
&=
\left\|
\sum_{t=1}^T q_t(x)\,\mathbf e_{\ell_t(x)}
\right\|_0
=
\sum_{t=1}^T \|q_t(x)\,\mathbf e_{\ell_t(x)}\|_0
\\
&=
\sum_{t=1}^T \mathbb I(q_t(x)\neq 0)
=
\|\boldsymbol q(x)\|_0
\le T,
\end{align*}
where the second equality follows from the fact that the supports of
$\mathbf e_{\ell_t(x)}$ are pairwise disjoint across $t$, since
$\ell_t(x)\neq \ell_{t'}(x)$ for any $t\neq t'$.
\end{proof}

\begin{lemma}[Constructive bilinear representation]
\label{lem:bilinear_rep}
Let $P_{\boldsymbol q,\boldsymbol w}$ be a SWLC proximity (Definition~\ref{def:swlcp}). Then, for all $x,x'\in\mathcal X$:
\begin{equation*}
P_{\boldsymbol q,\boldsymbol w}(x,x')
=
\langle
\boldsymbol{\phi}_{\boldsymbol{q}}(x),
\boldsymbol{\phi}_{\boldsymbol{w}}(x')
\rangle
=
\boldsymbol{\phi}_{\boldsymbol{q}}(x)^\top\boldsymbol{\phi}_{\boldsymbol{w}}(x').
\end{equation*}
\end{lemma}

\begin{proof}
See Appendix~\ref{app:bilinear_rep}.
\end{proof}

% \begin{proof}
% Fix $x,x'\in\mathcal X$. By Definition~\ref{def:weighted_leaf_representations},
% \begin{align*}
% \boldsymbol{\phi}_{\boldsymbol{q}}(x)^\top\boldsymbol{\phi}_{\boldsymbol{w}}(x')
% &=
% \left[\sum_{t=1}^T q_t(x)\,\mathbf e_{\ell_t(x)}\right]^\top\!
% \left[\sum_{s=1}^T w_s(x')\,\mathbf e_{\ell_s(x')}\right]
% \\
% &=
% \sum_{t=1}^T \sum_{s=1}^T
% q_t(x)\,w_s(x')\,
% \mathbf e_{\ell_t(x)}^\top\mathbf e_{\ell_s(x')}^{\phantom{^\top}}
% \\
% &=
% \sum_{t=1}^T
% q_t(x)\,w_t(x')\,
% \mathbf e_{\ell_t(x)}^\top\mathbf e_{\ell_t(x')}^{\phantom{^\top}}
% \\
% &=
% \sum_{t=1}^T
% q_t(x)\,w_t(x')\,
% \mathbb I(\ell_t(x)=\ell_t(x'))
% \\
% &=
% P_{\boldsymbol q,\boldsymbol w}(x,x'),
% \end{align*}
% where the third equality uses that $\mathbf e_{\ell_t(x)}^\top\mathbf e_{\ell_s(x')}=0$  since $\ell_t(x)\neq \ell_s(x)$ for $t\neq s$,   and the fourth equality uses $\mathbf e_a^\top \mathbf e_b^{\phantom{^\top}} = \mathbb I(a=b)$.
% \end{proof}

The bilinear form in Lemma~\ref{lem:bilinear_rep} formalizes the distinct functional roles of the two arguments. In asymmetric SWLC proximities such as RF-GAP~\cite{rhodes_geometry-_2023}, the weight assignments $\boldsymbol{q}$ and $\boldsymbol{w}$ correspond to the roles of \emph{query} and \emph{reference}, respectively. The query weight $\boldsymbol{q}$ typically captures tree-level importance or sample-specific relevance, while the reference weight $\boldsymbol{w}$ captures leaf-level influence (e.g., density normalization). For out-of-sample (OOS) extension, an unseen sample $x$ is by convention treated as the query, and its proximity to the training set is computed by projecting it via $\boldsymbol{\phi}_{\boldsymbol{q}}(x)$ and measuring its similarity against the gallery of reference representations $\{\boldsymbol{\phi}_{\boldsymbol{w}}(x_i)\}_{i=1}^N$ via the dot product in the leaf space.

While this latent dot-product structure has been observed in prior work, existing results are limited to a few specific tree-ensemble proximity definitions~\cite{vu2025autoencoding, whitmore_explicating_2018}. Here, we generalize this view to the full SWLC family, including both symmetric and asymmetric constructions, where the similarity takes the form of a bilinear product that, in general, involves two distinct leaf-incidence maps, $\boldsymbol{\phi}_{\boldsymbol{q}}$ and $\boldsymbol{\phi}_{\boldsymbol{w}}$. The previously studied symmetric cases therefore correspond to a strict subset of SWLC, whereas our framework also covers more general weighted and directed proximity definitions. The proposition below is our central result, as it converts this bilinear form into a practical finite-sample sparse matrix factorization.

\begin{proposition}[Exact finite-sample sparse factorization]\label{prop:finite_factorization}
Let $P_{\boldsymbol q,\boldsymbol w}$ be a SWLC proximity (Definition~\ref{def:swlcp}).
Define matrices $Q,W\in\mathbb R^{L\times N}_{\geq 0}$ by stacking the weighted leaf-incidence vectors,
\begin{equation*}
Q_{:i}:=\boldsymbol{\phi}_{\boldsymbol{q}}(x_i)\in\mathbb R^{L}_{\geq 0},
\qquad
W_{:j}:=\boldsymbol{\phi}_{\boldsymbol{w}}(x_j)\in\mathbb R^{L}_{\geq 0}.
\end{equation*}
Then the training proximity matrix $P=[P_{\boldsymbol q,\boldsymbol w}(x_i,x_j)]\in \mathbb{R}^{N\times N}$ admits the exact factorization
\begin{equation*}
P = Q^\top W.
\end{equation*}
Moreover, each column $i$ of $Q$ $(W)$ has exactly $\|\boldsymbol q(x_i)\|_0\leq T$ $(\|\boldsymbol w(x_j)\|_0\leq T)$ nonzeros.
\end{proposition}
\begin{proof}
By Lemma~\ref{lem:bilinear_rep}, for all $i,j$,
\[
P_{ij}
=
P_{\boldsymbol q,\boldsymbol w}(x_i,x_j)
=
\boldsymbol{\phi}_{\boldsymbol{q}}(x_i)^\top \boldsymbol{\phi}_{\boldsymbol{w}}(x_j)
=
Q_{:i}^\top W_{:j}.
\]
Thus $(Q^\top W)_{ij}=P_{ij}$. The column-sparsity statement follows from Lemma~\ref{cor:l0_weighted} applied to $\boldsymbol{\phi}_{\boldsymbol{q}}$ and $\boldsymbol{\phi}_{\boldsymbol{w}}$.
\end{proof}

\begin{corollary}[Gram kernel under identical weight assignment]
\label{cor:psd_q_equals_w}
Assume $\boldsymbol q=\boldsymbol w$ and let
$Q\in\mathbb R^{L\times N}_{\geq 0}$ be defined by
$Q_{:i}=\boldsymbol{\phi}_{\boldsymbol q}(x_i)$.
Then the SWLC proximity matrix reduces to a Gram kernel:
\[
P = Q^\top Q = W^\top W,
\]
and is therefore symmetric positive semidefinite. Moreover, among all nonnegative weighted leaf-incidence maps defined over the
fixed forest leaf coordinates, the reproducing feature map
$\boldsymbol{\phi}_{\boldsymbol q}$ is unique.
\end{corollary}

\begin{proof}
If $\boldsymbol q=\boldsymbol w$, then both arguments of the proximity share
the same feature map $\boldsymbol{\phi}_{\boldsymbol q}$. By
Proposition~\ref{prop:finite_factorization}, we have $Q=W$ by construction, and
\[
P = Q^\top W = Q^\top Q = W^\top W.
\]
Thus $P$ is a Gram matrix induced by the embedding
$
x \mapsto \boldsymbol{\phi}_{\boldsymbol q}(x)
$,
which implies symmetry and positive semidefiniteness, since
$
c^\top P c = \|Q c\|_2^2 \ge 0
$
for any $c\in\mathbb R^N$. For uniqueness, let
$
\phi_{\boldsymbol q}(x)
$
be another nonnegative weighted leaf-incidence map satisfying the reproducing
property. Then, for any $x$, $\langle
\boldsymbol{\phi}_{\boldsymbol{r}}(x),
\boldsymbol{\phi}_{\boldsymbol{r}}(x)
\rangle = P(x,x) = \langle
\boldsymbol{\phi}_{\boldsymbol{q}}(x),
\boldsymbol{\phi}_{\boldsymbol{q}}(x)
\rangle$ which implies 
$
r_t(x)^2=q_t(x)^2$ for all $t\in\{1,\ldots,T\}.
$ Since $r_t(x),q_t(x)\geq 0$, it follows that $
r_t(x)=q_t(x)$.
\end{proof}

\begin{figure}[!ht] %% placed here for LaTeX float positioning
    \centering
     \includegraphics[width = 0.92\columnwidth]{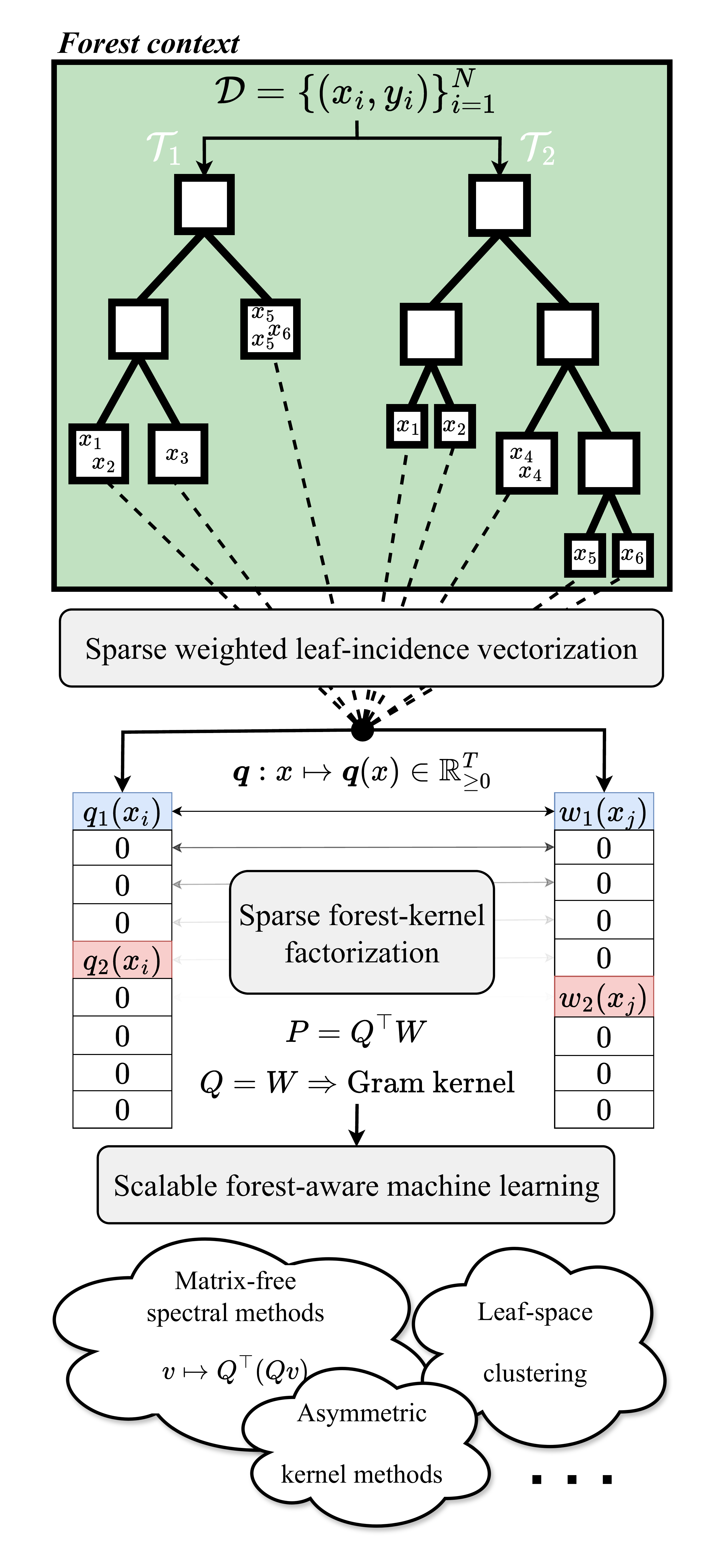}
    \caption{\textbf{Overview of the proposed sparse leaf-incidence factorization.} Forest proximities are represented as $P=Q^\top W$ using sparse weighted leaf-incidence maps, enabling scalable proximity computation and downstream forest-aware learning, including matrix-free spectral methods and asymmetric kernel methods, without explicitly constructing the dense $N\times N$ proximity matrix.}
    \label{fig:framework}
\end{figure}

Figure~\ref{fig:framework} summarizes the proposed sparse leaf-incidence factorization and its main computational implications. Beyond enabling scalable proximity computation, this decomposition is also a structural result: it makes the feature maps underlying SWLC proximities explicit and shows that these proximities can be interpreted as inner products between leaf-based representations. This perspective clarifies why SWLC proximities integrate naturally with kernel methods such as kernel PCA and kernel support vector machines (SVM), rather than behaving as black-box similarities. It also explains why kernel PCA could be successfully applied to forest-based proximity matrices in prior work~\cite{rhodesGainingBiologicalInsights2026, rhodes_supervised_2023}.

In particular, in the symmetric case, the singular value decomposition (SVD) theorem shows that the spectral structure of the full proximity matrix $P$ can be recovered directly from the explicit leaf-incidence matrix $Q$ \cite{horn2012matrix}. As a result, spectral methods formulated on $P$ can be implemented equivalently on $Q$, without ever forming the dense matrix $P$. This is particularly attractive computationally, since it allows one to exploit sparse linear algebra directly in leaf space~\cite{lehoucq1998arpack, halko2009finding, virtanen_scipy_2020}.

%This perspective also extends beyond the symmetric setting. 
Kernel-style methods have also been generalized to asymmetric kernels~\cite{he_learning_2023, suykens_svd_2016, tao_learning_2024}, which can capture richer directional relationships. Proposition~\ref{prop:finite_factorization} thus provides a principled foundation for developing  general supervised kernel methods based on directed forest proximities.

\subsection{Complexity analysis.}\label{subsec:complexity}
Computing $P$ from scratch involves four stages: (i) training the ensemble $\mathcal{T}$, (ii) computing the metadata $\theta$ required by the weight assignments $(\boldsymbol{q},\boldsymbol{w})$, (iii) constructing the sparse factors $Q$ and $W$, and (iv) evaluating the product $Q^\top W$. For standard tree ensemble methods, if
$
\bar h := \frac{1}{T}\sum_{t=1}^T h_t
$ denotes the average tree height, training and routing scale as $\mathcal O(NT\bar h)$ under common implementations~\cite{louppe_understanding_2015, hassine_important_2019}. Likewise, when the metadata is sample-wise, tree-wise, or leaf-wise, as in most proximity definitions, it can typically be obtained by routing followed by local leaf aggregation, with cost at most $\mathcal O(NT\bar h)+\mathcal O(NT)$. Thus, these preprocessing steps do not introduce a quadratic scaling in $N$.

The main remaining bottleneck is the proximity computation itself. A naive implementation would still require evaluating all $N^2$ sample pairs across trees, yielding $\mathcal O(N^2T)$ complexity. We now show that the sparse product $Q^\top W$ avoids this quadratic cost by exploiting the sparsity patterns of $Q$ and $W$, so that computation is restricted to sample pairs that collide in at least one leaf.

\paragraph{Building $Q$ and $W$.}
Although $Q$ and $W$ are mathematically $L\times N$, Proposition~\ref{prop:finite_factorization} implies only $\mathcal O(NT)$ memory is required in compressed sparse format. Constructing them requires routing each sample through each tree. Thus, this stage costs
$
\mathcal O(NT\bar h).
$

\paragraph{Sparse product $Q^\top W$.}
The computational gain comes from the fact that sparse multiplication does not evaluate all $N^2$ sample pairs densely. For any pair $(i,j)$,
$
P_{ij}
=
\boldsymbol{\phi}_{\boldsymbol q}(x_i)^\top
\boldsymbol{\phi}_{\boldsymbol w}(x_j)
$
is non-zero only if the two leaf-incidence vectors share at least one active coordinate, that is, only if $x_i$ and $x_j$ fall in the same leaf of at least one tree. This is precisely the mechanism exploited by \texttt{scipy}'s sparse routines~\cite{virtanen_scipy_2020}: the product is accumulated only through shared non-zero column indices, rather than by checking all $N^2$ pairs explicitly.

To quantify this, fix a sample $x_i$ and a tree $t$. Instead of comparing $x_i$ to all $N$ samples, tree $t$ only induces interactions with the samples that fall in the same leaf as $x_i$. If
$
n_{t,\ell_t(x_i)}
$
denotes the number of samples in that leaf, then the work associated with sample $x_i$ across all trees is proportional to
$
\sum_{t=1}^T n_{t,\ell_t(x_i)}.
$
Summing over all samples yields the global sparse multiplication cost
$
\mathcal O\!\left(\sum_{i=1}^N\sum_{t=1}^T n_{t,\ell_t(x_i)}\right)
=
\mathcal O(NT\bar\lambda),
$
where
$$
\bar\lambda
:=
\frac{1}{NT}\sum_{i=1}^N\sum_{t=1}^T n_{t,\ell_t(x_i)}
$$
is the average number of same-leaf interactions per sample and per tree.

The combined routing and sparse product cost are:
\begin{align*}
\textsc{time}
\quad &:\quad
\mathcal O(NT\bar h)+\mathcal O(NT\bar\lambda)
=
\mathcal O\!\bigl(NT(\bar h+\bar\lambda)\bigr),
\\
\textsc{space}
\quad &:\quad
\mathcal O(NT)+\mathcal O(NT\bar\lambda)
=
\mathcal O\!\bigl(NT(1+\bar\lambda)\bigr).
\end{align*}
Thus, the complexity of the factorized computation is governed by $N$, $T$, the average tree height $\bar h$, and the average leaf-collision factor $\bar\lambda$. In standard regimes where $T\ll N$, $\bar h\approx \log N$ for balanced trees, and $\bar\lambda\ll N$ for sufficiently deep trees, the full pipeline is near-linear in the sample size (or at least sub-quadratic).

\begin{remark}\label{rem:tight_complexity_bound}
These bounds can be tightened for specific sparse weight assignments, but we omit this refinement for simplicity, as our focus is the scaling behavior of the full SWLC family. See Section~\ref{subsec:scaling_results} for an empirical comparison of different weighting schemes.
\end{remark}

\begin{remark}
    The same reasoning applies to OOS extension. If $Q_{\mathrm{new}}\in\mathbb R^{N_{\mathrm{new}}\times L}$ denotes the sparse leaf-incidence matrix of $N_{\mathrm{new}}$ unseen samples, then computing cross-proximities via
$
Q_{\mathrm{new}}^\top W
$
costs
$
\mathcal O(N_{\mathrm{new}}T\bar h)
\;+\;
\mathcal O(N_{\mathrm{new}}T\bar\lambda_{\mathrm{ext}}),
$
where the first term accounts for routing the new samples through the ensemble and building $Q_{\mathrm{new}}$, and the second term accounts for the sparse multiplication. Here,
$
\bar\lambda_{\mathrm{ext}}
$
denotes the average number of reference samples sharing a leaf with a queried sample, per tree. Hence, OOS proximity extension is also near-linear in the number of queried samples when $\bar h$ and $\bar\lambda_{\mathrm{ext}}$ are bounded or slowly growing.
\end{remark}

\section{ Results.}\label{sec:results}

We implemented our sparse forest kernel in a user-friendly Python interface built on top of standard \texttt{scikit-learn}~\cite{scikit-learn} forest estimators, which made it easy to integrate with existing workflows while keeping the API familiar. All experiments were run on a CPU compute node. See Appendices~\ref{app:implementation} and~\ref{app:env} for details on the computing environment and implementation. In the experiments below, we consider 11 medium- to large-scale labeled datasets spanning diverse domains, feature dimensions, and numbers of classes, to illustrate the relevance of our sparse kernel decomposition across a broad range of settings. A summary of the datasets is provided in Appendix~\ref{app:datasets}.

\subsection{Asymptotic separability of OOB proximity.}\label{subsec:empirical_oob_separability}

Proposition~\ref{prop:swlc} shows that standard OOB proximities are non-separable. We propose an alternative, $\tilde{P}_{\mathrm{oob}}$, which achieves separability by normalizing leaf collisions with the proxy $S(x)S(x')/T$ instead of the shared OOB tree count $S(x,x')$. This formulation enables scalable computation via sparse kernel decomposition and explicit leaf coordinates while remaining asymptotically equivalent to the standard definition as $T$ and $N$ increase (see Appendix~\ref{app:separable_oob}).

\begin{figure}[!ht] %% placed here for LaTeX float positioning
    \centering
     \includegraphics[width = 0.85\columnwidth]{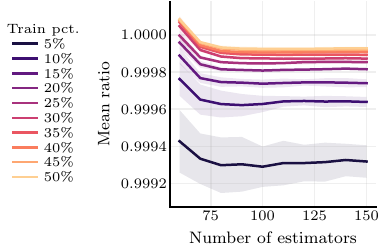}
    \caption{Mean ratio $R(x,x') = \frac{S(x,x')}{S(x)S(x')/T}$ with standard deviations on SignMNIST (A–K) vs. number of trees $T$. Curves represent training sizes $N$ from 5\% to 50\%. As $T$ and $N$ increase, the ratio converges toward 1 from below, validating the asymptotic separability of Proposition~\ref{prop:oob_shared_count} and the fidelity of the $\tilde{P}_{\mathrm{oob}}$ proxy.}
    \label{fig:separable_oob}
\end{figure}

We empirically validate this asymptotic separability on SignMNIST~\cite{sign_language_mnist}. We evaluate the average ratio $R(x,x') = \frac{S(x,x')}{S(x)S(x')/T}$ across distinct samples where $S(x,x') > 0$, varying training size $N$ (5\%--50\%) and forest size $T$ (60--150). Results are shown in Fig.~\ref{fig:separable_oob}, and support our Proposition~\ref{prop:oob_shared_count}: the ratio stabilizes as $T$ grows, and approaches 1 from below as $N$ increases. In standard regimes, the approximation error is negligible, providing a faithful and scalable alternative to standard OOB formulations (Fig.~\ref{fig:general_scaling}).

\subsection{Empirical scaling of sparse forest kernel computation.}\label{subsec:scaling_results}

We empirically validate the near-linear time and memory scaling in the number of samples established in Section~\ref{subsec:complexity}, across a range of settings.
We use standard RF training as implemented in \texttt{scikit-learn}’s \texttt{RandomForestClassifier}, together with RF-GAP weight assignments. For evaluation, each dataset is split into train and test sets using a 10\% stratified split (or predefined splits when available).

\begin{figure}[!ht] %% placed here for LaTeX float positioning
    \centering
     \includegraphics[width = \columnwidth]{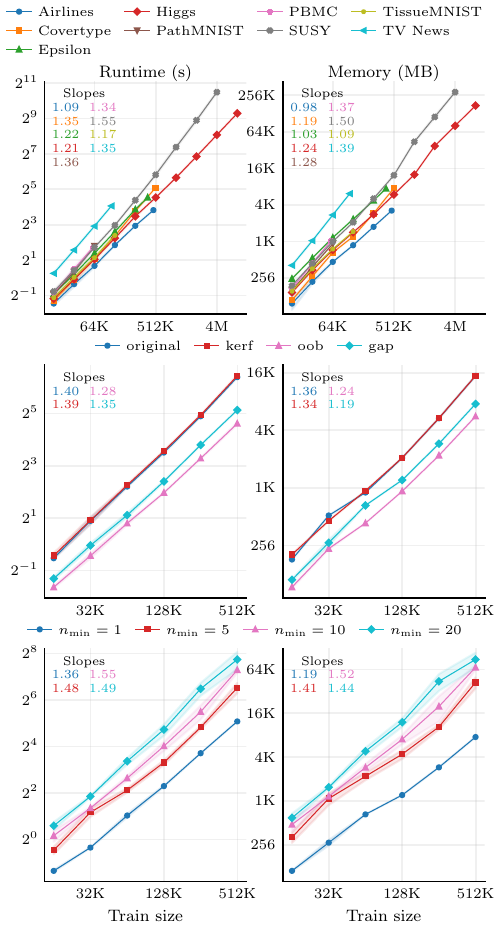}
    
        \caption{Log--log runtime and memory scaling of exact kernel computation with sample size, across datasets, proximity methods, and minimum leaf sizes. Shaded bands show standard deviations, and slopes are from fitted linear regressions. Overall, computation scales near-linearly with sample size, with slopes well below $2$. RF-GAP and our separable OOB kernel scale more favorably, consistent with the extra sparsity induced by their weighting schemes. Larger minimum leaf sizes degrade scaling, but remain practical and robust across a reasonable range of values.}

    \label{fig:general_scaling}
\end{figure}

We analyze the time and memory required to compute the full kernel as the number of training samples increases, under three axes of variation: (i) across nine datasets of varying scale, including the Higgs dataset~\cite{baldi2014searching} with up to 10 million samples (Fig.~\ref{fig:general_scaling}, top), (ii) across four proximity definitions—original, KeRF, our \textit{separable} OOB, and RF-GAP (Fig.~\ref{fig:general_scaling}, middle); and (iii) across the minimum leaf size parameter $n_{\min}$ (Fig.~\ref{fig:general_scaling}, bottom). We performed (i) and (ii) on Covertype. Additional ablations, including experiments with constrained tree depth and additional datasets, are provided in Appendix~\ref{app:full_ablation}. We also report, in Appendix~\ref{app:sanity_check_kernel_acc}, the test accuracy of proximity-weighted predictions as a sanity check to assess the usefulness of the mined kernels for downstream classification tasks.

Reported runtime and memory correspond to the cost of constructing and storing the cached metadata, query maps, and the resulting sparse kernel. Forest training is excluded, as it is independent of our decomposition and does not dominate computational cost.

Overall, both runtime and memory exhibit near-linear scaling (with slopes well below 2 and often close to 1), in agreement with our theoretical analysis. On the Higgs dataset, computing the full kernel on a 1M-sample subset requires under one minute and less than 16GB of memory, making the approach practical on standard hardware. Across proximity definitions, original and KeRF show similar scaling, as both rely on full leaf collisions. In contrast, the OOB kernel is the most scalable, since it only considers OOB–OOB interactions, while GAP lies in between due to its OOB–in-bag structure. This aligns with Remark~\ref{rem:tight_complexity_bound}, which suggests that tighter bounds can be obtained by accounting for sparsity induced by specific weighting schemes. Finally, increasing the minimum leaf size degrades performance compared to unconstrained forests, but near-linear scaling is preserved over a broad range of $n_{\min}$, making the approach robust to practical hyperparameter tuning.

\subsection{Manifold learning on leaf coordinates.}~\label{sec:manifold}
Beyond enabling scalable full kernel computation, our explicit kernel decomposition opens another promising direction: injecting the structure learned by forest models into downstream manifold learning methods that typically operate directly on the  feature space.

Our leaf maps live in a space of dimension $L$, which in the worst case can be as large as $NT$. Although each vector is only $T$-sparse, most manifold learning pipelines expect dense inputs, and therefore do not naturally handle sparse leaf coordinates without an intermediate DR step. The most common choice is PCA, which is  often used as a preprocessing step for nonlinear DR methods. Fortunately, the \texttt{scikit-learn} implementation of PCA provides an ARPACK solver~\cite{lehoucq1998arpack} that can operate on sparse inputs through linear operators, without explicitly centering the full matrix in leaf space. Since the runtime of many manifold learning methods is dominated by neighbor search and graph construction, computing Leaf PCA coordinates is not typically the bottleneck. This makes leaf coordinates a practical way to incorporate forest-based supervision into existing pipelines.

Our decomposition also provides a useful conceptual point of view. When standard forest proximities are used as precomputed kernels, the resulting DR pipeline implicitly restricts to linear leaf-collision relationships. In contrast, our explicit leaf coordinates expose the same structure in a vector representation, making it possible to go beyond standard kernel-based workflows and apply more general nonlinear manifold learning methods directly in the leaf space. We explore this here. 

\begin{figure*}[!ht] %% placed here for LaTeX float positioning
    \centering
     \includegraphics[width = \textwidth]{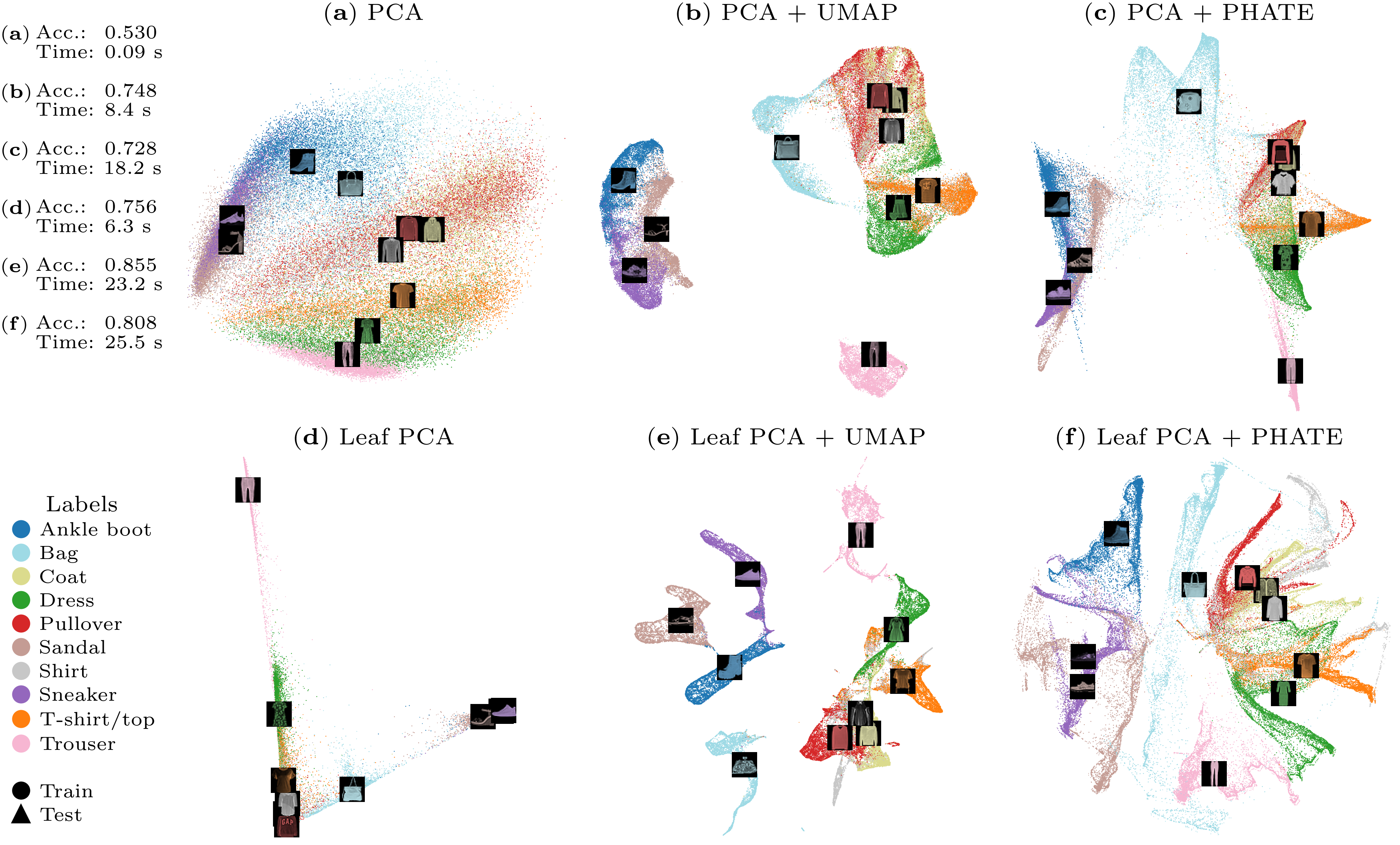}
    \caption{Two-dimensional train and test embeddings of FashionMNIST samples obtained from raw flattened pixel features using (a) PCA, (b) PCA followed by UMAP, and (c) PCA followed by PHATE, compared with the same pipelines applied to sparse forest leaf coordinates in (d–f). The top-left of each panel reports runtime and test $k$-NN accuracy. Images show class-wise medoids in the embedding space. Overall, sparse leaf coordinates provide a scalable, denoised, task-aware representation for existing manifold learning pipelines.}
    \label{fig:leaf_manifold_learning}
\end{figure*}

We consider the FashionMNIST dataset~\cite{xiao_fashion-mnist_2017}, with 60,000 training samples and 10,000 test samples. We study two families of DR pipelines: PCA alone, PCA followed by UMAP~\cite{McInnes2018}, and PCA followed by PHATE~\cite{moon2019visualizing}, with 50 principal components used as preprocessing for UMAP and PHATE. The second family applies the same pipelines to sparse leaf coordinates instead of raw image features. For simplicity, we use KeRF leaf coordinates, which ensure symmetry for the PCA step. We otherwise use default parameters for all methods, except that we set $k=30$ for UMAP and PHATE to ensure graph connectedness during $k$-NN construction. Each pipeline is trained on the raw training images, and both train and test samples are embedded in two dimensions. In Fig.~\ref{fig:leaf_manifold_learning}, we report the runtime of each pipeline and the average test embedding $k$-NN accuracy for $k=5,10,20$, using the training embedding as reference. We also display the corresponding classwise medoid images on each embedding.

All DR pipelines benefited from leaf coordinates in terms of class separability, especially PCA. This is expected: raw PCA is purely linear, whereas Leaf PCA introduces nonlinearity through the leaf-incidence maps, making it analogous to  kernel PCA. While raw PCA produces a noisy embedding, Leaf PCA reveals two dominant directions that are both class-aware and semantically meaningful: a horizontal direction related to accessories, with bags and, farther away, shoes, and a vertical direction related to broader clothing types such as dresses and trousers. This makes Leaf PCA a strong tool for visualizing forest-guided global structure.

Leaf UMAP and Leaf PHATE act more as supervised refinements of their unsupervised counterparts. In raw UMAP and raw PHATE, classes such as coat, shirt, and pullover, or ankle boot, sandal, and sneaker, remain partly mixed. Their leaf-based versions separate these groups more clearly without distorting their relationships. Leaf UMAP emphasizes classwise cluster separation, while Leaf PHATE highlights connections between clusters and class-aware trajectories. Leaf PHATE also shows a progressive transition among coats, shirts, and pullovers in the top-right part of the embedding. This is meaningful, since these classes can be viewed as related variants of a broader clothing type. From a computational perspective, the runtime of Leaf UMAP and Leaf PHATE is dominated by the UMAP or PHATE stage, while the Leaf PCA step is negligible by comparison. We provide a similar visual comparison on SignMNIST (A--K) in Appendix~\ref{app:sign_mnist}. Overall, the exposed leaf space provides a scalable coordinate system that integrates supervision smoothly into existing  DR pipelines.

\section{Discussion.}

Forest proximities have long been used as supervised similarities for visualization, prediction, matching, and explanation, but their broader use has been limited by their quadratic scalability. 
Our work addresses this scalability. On the conceptual side, we introduce the SWLC framework, which unifies a broad class of forest proximities through a common leaf-collision form, making their feature-space structure explicit. In symmetric cases, this yields a sparse Gram representation and a direct kernel view. In asymmetric cases, it yields a bilinear leaf-space view that also covers directed proximities. Computationally, this leads to an exact sparse factorization that avoids explicit pairwise comparison across all samples. Our experiments show that this gives strong gains in runtime and memory, and that these gains remain stable across datasets, ensemble types, kernel variants, and forest depth settings.

Because the factorization exposes explicit sparse leaf-incidence representations, spectral methods can operate directly on the leaf maps instead of on the full proximity matrix. In the symmetric case, this gives the same uncentered spectral structure as the kernel route, while fitting naturally within standard sparse linear algebra pipelines. Our results confirm that this direct leaf route preserves the relevant geometry and reduces computational cost.

A limitation is that our full kernel complexity bounds rely on computation being restricted to leaf-colliding samples. This is favorable in the usual regime of reasonably deep trees, but may weaken when trees are too shallow, and leaves become large. Still, our empirical results suggest that the near-linear behavior is robust in a broad practical regime.

The framework opens several natural directions. One is to enrich existing separable proximities, e.g., by adding leaf-quality statistics such as impurity measures, or by combining ideas from different ensemble paradigms such as boosting and bagging~\cite{ganjisaffar_bagging_2011}. Another is to move from fixed weighting rules to learned ones while keeping the forest topology fixed.  This would yield sample-dependent forest kernels that remain faithful to the supervised geometry induced by the ensemble, while preserving  sparsity. In this sense, our framework also points toward forest-based kernel learning, including tree reweighting schemes related to multiple kernel learning~\cite{sonnenburg2006large} and differentiable optimization on top of fixed tree structures~\cite{hazimeh_tree_2020}.
Finally, and perhaps the most promising direction, is to pursue our preliminary results in Section~\ref{sec:manifold} on the integration of the leaf coordinates into downstream manifold learning.

On the implementation side, we plan to broaden the range of supported ensemble contexts and weighting schemes, to support user-defined forest kernels, and to accelerate the full pipeline further through GPU support for training, metadata extraction, and sparse proximity computation. We expect this work to move forest proximities from a set of specialized algorithms to a general, scalable, and explicit kernel framework.

\section*{Acknowledgements}
\noindent This research was enabled in part by compute resources provided by the Department of Mathematics and Statistics at the Université de Montréal. It was supported in part by the Ministry of Health and Social Services (Quebec) in collaboration with the Centre intégré de santé et de services sociaux Centre-Sud-de-l'Île-de-Montréal [A.A.], the Outstanding PhD Candidate Scholarship from the Institut des sciences mathématiques [A.A.], Canada CIFAR AI Chair [G.W.], NSERC Discovery grant 03267 [G.W.], NIH grant R01GM135929 [G.W.], NSF grant DMS-2327211 [G.W.], NSF grant 221232 [K.M], NIH grant 1R15HL168697 [K.M.], and the IVADO Visiting Scholar program [K.M]. The authors are solely responsible for the content of this work, and the views expressed in it do not necessarily reflect those of the funding agencies. The authors sincerely thank the anonymous reviewers, the Area Chairs, and the Program Chairs for their constructive feedback and their efforts in organizing the review process. We also gratefully acknowledge the assistance of the volunteers who prepared and maintained the SIAM Proceedings \LaTeX{} macros and example files.

\clearpage

\bibliographystyle{siamplain}

\bibliography{references_zotero}

\appendix

\section{Extended literature review on forest proximities.}\label{app:overview_extended}

\subsection{Proximities based on leaf collisions.}

The original forest proximity, introduced by Breiman~\cite{breiman_random_2001,breiman_manual_2003}, defines similarity between two samples as the proportion of trees in which they fall into the same terminal leaf. It is the most direct and canonical notion of leaf-collision similarity: two points are considered close when the ensemble repeatedly groups them together through learning a feature space partition. Despite its simplicity, the original proximity remains widely used and can be highly competitive in practice in specific contexts~\cite{farokhi_advancing_2024}.

A normalized variant from the kernel random forest (KeRF) literature~\cite{scornet2016random} refines the original definition by downweighting collisions that occur in large leaves. The intuition is that sharing a very large terminal cell provides a weaker notion of task-aware similarity than sharing a small, exclusive leaf. This leaf-mass normalization preserves symmetry and yields a kernel with useful stochastic properties, making it particularly attractive for downstream diffusion-based methods~\cite{vu2025autoencoding}.

Several variants modify collision counting using additional information from the training procedure, such as bootstrap aggregation. The original RF proximities can be biased because they ignore each observation's bootstrap status, that is, whether it was in-bag or OOB for a given tree. %In standard definitions, in-bag and OOB co-occurrences are counted and weighted equally. 
Since trees are typically grown to purity, in-bag samples from different classes almost always end up in different terminal leaves, which can artificially amplify class separation in the resulting proximity matrix. 
OOB proximity~\cite{liaw2002classification,Hastie2009} addresses this effect by counting leaf collisions only in trees for which both samples are simultaneously OOB. This definition is used in the \texttt{randomForest} package~\cite{liaw2002classification} for the R programming language~\cite{team2020ra}.

Along this line, RF-GAP~\cite{rhodes_geometry-_2023} combines OOB querying with leaf-mass normalization and replaces binary indicators with the full per-tree in-bag multiplicity of each training sample. The resulting affinities are more closely aligned with the geometry that the forest learns for prediction. In particular, RF-GAP induces a proximity-weighted predictor that exactly recovers the RF OOB predictions~\cite{rhodes_geometry-_2023}. It has been shown to outperform OOB proximities and several classical alternatives in tasks such as classification, regression, imputation, and outlier detection~\cite{rhodes_geometry-_2023}. A drawback is that RF-GAP is generally asymmetric, making it incompatible with standard downstream methods requiring symmetry, although a symmetric variant can be obtained by averaging the two directions while retaining much of the practical benefit of the directed formulation~\cite{rhodes_geometry-_2023,rhodesGainingBiologicalInsights2026,aumon2025random}.

Other RF-based variants also retain the same leaf-collision principle but reweight collisions using additional sample-level information. RFProxIH is an adaptation of the RFDisIH dissimilarity~\cite{cao_novel_2021} for multi-view classification. Rather than recovering the forest’s own geometry, it uses the forest partition as a scaffold and re-weights leaf collisions using an external notion of instance hardness (IH). It is therefore best viewed as a task-specific reweighting of leaf collisions, rather than as a forest-native interpretability tool. The authors reported strong classification performance for RFDisIH relative to other multi-view baselines, but did not compare it to other random-forest proximity definitions in common proximity-driven applications such as visualization or imputation.

% GBTs
The literature on proximities for GBTs is more limited, but has gained traction recently. Tan et al.~\cite{tan_tree_2020} extended the original leaf-collision proximity to boosted ensembles by weighting each tree’s collision indicator by that tree’s contribution to the additive model. They showed that this tree-weighted proximity can outperform Euclidean and RF-induced distances for prototype selection with $k$-medoids. More recently, the Additive eXplanations with Instance Loadings (AXIL~\cite{geertsema_instance-based_2023}) framework derives a supervised proximity for GBTs by rewriting the model prediction as an additive combination of training targets and interpreting the resulting weights as a similarity measure. Unlike purely heuristic tree-weighting schemes, AXIL is closer in spirit to RF-GAP: it incorporates leaf-size normalization so that the induced weights better explain the numerical output of the fitted regression model, and it discusses extensions to classification as future work. However, unlike classical forest proximities, AXIL weights are not constrained to be nonnegative and can become signed through the recursive residual corrections of boosting. Consequently, they are arguably better interpreted as instance-wise influence coefficients than as a conventional geometric affinity or proximity measure. The authors further show that these AXIL weights provide useful, complementary instance-based explanations in a smoking-prevalence application, alongside feature-based tools such as SHAP~\cite{lundberg_unified_2017} and LIME~\cite{ribeiro_why_2016}.

\subsection{Other forest proximities.}

Proximity-based kernel (PBK;~\cite{englund_novel_2012}) departs from strict leaf collisions by incorporating the tree topology: instead of assigning similarity only when two samples land in the same leaf, it measures how far apart they are within a tree. PBK defines the proximity between two observations as the average, across trees, of an exponentially decaying function of the number of branches separating them. This yields a nonzero similarity even between points in different leaves, which can capture finer graded relationships than the original proximity. The authors report improved robustness over the original proximity for shallow forests when used with SVM classifiers. The tradeoff is that PBK introduces a decay hyperparameter and is substantially more expensive to compute, since it requires evaluating branch distances for many pairs; in practice, its computational cost scales poorly, and reported gains appear mostly confined to small ensembles.

Random Partitioning Kernel (RPK~\cite{davies_random_2014}) views tree-based kernels through the lens of random data partitions. Rather than relying primarily on terminal node co-occurrence, RPK samples a random tree height and partitions the data using that intermediate level of the tree, effectively defining similarity through co-membership in a randomly chosen coarse partition. The authors compare this RF kernel to standard kernels (e.g., linear, RBF) using a log-likelihood based evaluation, reporting favorable performance in many cases. They illustrate the induced structure with 2D visualizations through principal component analysis (PCA); however, they do not position the method as a proximity definition alongside existing forest proximities, and an implementation is not publicly available. Both PBK and RPK are therefore less commonly used in the proximity literature and, importantly, they are not strictly leaf collision-based. While they are interesting in their own right, we do not align our framework with them here, and leave extensions in this direction for future work.

\section{Forest proximities as SWLC kernels.}
\label{app:swlcp_proofs}

In this section, we show that several common forest proximity measures can be expressed as instances of the Separable Weighted Leaf-Collision (SWLC) family defined in Definition~\ref{def:swlcp}. For each method, we specify an ensemble context $(\mathcal T,\theta)$ and a corresponding weight assignment $(\boldsymbol q,\boldsymbol w)$ that recovers the target proximity. While the ensemble context is fixed for each method, the associated weight assignment is generally not unique, and the same proximity can often be recovered from multiple valid choices of $(\boldsymbol q,\boldsymbol w)$. For symmetric SWLC proximities, however, we present identical weight assignments for both arguments; this is a sufficient condition for symmetry and yields the unique reproducing weighted leaf-incidence representation within the class of nonnegative weighted leaf-incidence maps (Corollary~\ref{cor:psd_q_equals_w}).

\subsection{Original proximities.}

Let $\mathcal T$ be a trained decision forest with $T$ trees. The original proximity (introduced in the context of RFs) depends only on the ensemble topology ($\theta=\emptyset$). It is defined by
\[
P_{\text{original}}(x,x')
=
\frac{1}{T}\sum_{t=1}^T
\mathbb{I}\!\bigl(\ell_t(x)=\ell_t(x')\bigr).
\]
This is an instance of the SWLC form with weights
\[
q_t(x)=w_t(x)=\frac{1}{\sqrt{T}}, \qquad t=1,\ldots,T.
\]
In particular, $P_{\text{original}}$ is a symmetric SWLC proximity since $q=w$.

\subsection{KeRF proximities.}

KeRF~\cite{scornet2016random} augments the ensemble with leaf-level statistics. $\mathcal{T}$ is a trained decision forest and $\theta = M \in \mathbb{R}^{L}$, where $M(j)$ denotes the mass (number of training samples) in leaf $j$. The proximity is defined as
\[
P_{\text{KeRF}}(x,x')
=
\frac{1}{T}\sum_{t=1}^T
\frac{\mathbb{I}\!\bigl(\ell_t(x)=\ell_t(x')\bigr)}{M\!\bigl(\ell_t(x)\bigr)}.
\]
Setting
\[
q_t(x)=w_t(x)=\frac{1}{\sqrt{T\,M(\ell_t(x))}},
\]
we recover $P_{\text{KeRF}}$. Indeed, whenever $\ell_t(x)=\ell_t(x')$, the denominator satisfies
$M(\ell_t(x)) = M(\ell_t(x'))$, so the product $q_t(x)\,w_t(x')$ yields $1/(T\,M(\ell_t(x)))$ on colliding leaves and $0$ otherwise. Hence, KeRF is a symmetric SWLC proximity.

\subsection{OOB proximities.}\label{subsec:oob}

OOB proximities incorporate bootstrap information. $\mathcal{T}$ is a trained decision forest via bootstrap aggregation, and $\theta$ includes (i) per-sample, per-tree OOB indicators $o_t(\cdot)\in\{0,1\}$, and (ii) the \emph{shared} OOB tree counts
\[
S(x,x')=\sum_{t=1}^T o_t(x)\,o_t(x'),
\]
which record the number of trees for which \emph{both} samples are OOB. The OOB proximity is defined as
\[
P_{\text{oob}}(x,x')
=
\frac{1}{S(x,x')}\sum_{t=1}^T
o_t(x)\,o_t(x')\,\mathbb{I}\!\bigl(\ell_t(x)=\ell_t(x')\bigr).
\]
The normalization factor \(S(x,x')\) depends jointly on the pair \((x,x')\) and, in general, does not admit a separable decomposition into independent functions of \(x\) and \(x'\). Thus, \(P_{\mathrm{oob}}\) does not belong to the SWLC family.

Nevertheless, OOB proximities remain symmetric, and the training OOB proximity matrix can be written as
\[
P = (Q^\top Q) \oslash S,
\]
where \(Q=W\) is the binary OOB leaf-incidence representation of the training data, \(S_{ij}=S(x_i,x_j)\), and \(\oslash\) denotes element-wise division.

In practice, we first compute \(Q^\top Q\), which remains near-linear in the number of samples due to the sparsity of the leaf-incidence representation (see Section~\ref{subsec:complexity}). The normalization matrix \(S\) is then evaluated only on pairs corresponding to nonzero entries of \(Q^\top Q\), so the full dense \(N\times N\) matrix is never materialized. This sparse normalization incurs an additional cost of
$
\mathcal O\!\left(T\,\mathrm{nnz}(P)\right),
$
since determining the shared OOB count for a nonzero pair requires inspecting the \(T\) tree-level OOB indicators. Although this introduces a modest overhead compared to separable proximities, the overall near-linear complexity is preserved. Moreover, the cost is partially offset by the OOB restriction itself, as interactions are only considered between samples sharing at least one OOB leaf collision.

When a separable formulation is required, for example to construct OOB variants of downstream matrix-free kernel methods, Proposition~\ref{prop:oob_shared_count} shows that the OOB proximity admits an asymptotically equivalent symmetric SWLC approximation with
$
q_t(x)=w_t(x)=\frac{\sqrt{T}}{S(x)},
$
where
$
S(x)=\sum_{t=1}^{T} o_t(x)
$
denotes the number of OOB trees associated with sample \(x\).

\subsection{GAP proximities.}

RF-GAP~\cite{rhodes_geometry-_2023} incorporates both leaf-mass normalization and sample-dependent reweighting based on bootstrap status. It is designed so that, when used as weights in a proximity-weighted predictor, the resulting estimator matches the forest's OOB predictions exactly. $\mathcal{T}$ is a trained decision forest via bootstrap aggregation, and $\theta$ contains (i) leaf in-bag masses $M_{\text{in-bag}}\in\mathbb{R}^{L}$, (ii) per-sample, per-tree OOB indicators $o_t(\cdot)$ and in-bag multiplicities $c_t(\cdot)$, and (iii) per-sample OOB tree counts $S(\cdot)$. The forest-agnostic GAP proximity is
\[
P_{\text{gap}}(x,x')
=
\frac{1}{S(x)}\sum_{t=1}^T
\frac{o_t(x)c_t(x')\,\mathbb{I}\!\bigl(\ell_t(x)=\ell_t(x')\bigr)}{M_{\text{in-bag}}(\ell_t(x'))}.
\]
This admits an SWLC representation by taking
\[
q_t(x)=\frac{o_t(x)}{S(x)},
\qquad
w_t(x)=\frac{c_t(x)}{M_{\text{in-bag}}(\ell_t(x))}.
\]
Therefore, $P_{\text{gap}}$ is a SWLC proximity, and is generally asymmetric.

\subsection{IH proximities.}

RFProxIH incorporates IH through the $k$-disagreeing neighbors (DN) score. $\mathcal T$ is a trained decision forest, and $\theta$ contains the collection of tree-dependent hardness functions $kDN_t(\cdot)$ defined relative to the labeled training set $\mathcal{D}=\{(x_i,y_i)\}_{i=1}^N$. The proximity is
\[
P_{\text{ih}}(x,x')
=
\frac{1}{T}\sum_{t=1}^T
\bigl(1-kDN_t(x')\bigr)\,
\mathbb{I}\!\bigl(\ell_t(x)=\ell_t(x')\bigr).
\]
Here,
\[
kDN_t(x)
=
\frac{1}{k}
\left|
\left\{\, j : x_j \in kNN_t(x)\subset X,\ y_j \neq y \,\right\}
\right|,
\]
where $kNN_t(x)$ denotes the set of $k$ nearest neighbors of $x$ computed in the tree-dependent feature subspace induced by the splitting variables along the decision path leading to the relevant leaf in tree $t$. This can be written in SWLC form by choosing
\[
q_t(x)=\frac{1}{T},
\qquad
w_t(x)=1-kDN_t(x).
\]
Therefore, $P_{\text{ih}}$ is an SWLC proximity and is generally asymmetric. A symmetric variant can be obtained, for example, by distributing the hardness factor across both arguments (e.g., using a symmetric combination of $kDN_t(x)$ and $kDN_t(x')$).

\subsection{Boosted proximities.}

In GBTs, each tree $t$ is associated with a nonnegative weight $w_t$ reflecting its contribution to the additive model. $\mathcal{T}$ is a trained sequential forest via gradient boosting, and $\theta=\{w_t\}_{t=1}^T$. The corresponding proximity~\cite{tan_tree_2020} is
\[
P_{\text{boosted}}(x,x')
=
\frac{\sum_{t=1}^T w_t\,\mathbb{I}\!\bigl(\ell_t(x)=\ell_t(x')\bigr)}
{\sum_{s=1}^T w_s}.
\]
This admits an SWLC representation by setting
\[
q_t(x)=w_t(x)=\sqrt{\frac{w_t}{\sum_{s=1}^T w_s}}.
\]
Therefore, $P_{\text{boosted}}$ is a symmetric SWLC proximity, since it uses identical weight assignments ($q=w$).

\section{Justification of Lemma~\ref{lem:bilinear_rep}.}\label{app:bilinear_rep}

\begin{proof}
Fix $x,x'\in\mathcal X$. By Definition~\ref{def:weighted_leaf_representations},
\begin{align*}
\boldsymbol{\phi}_{\boldsymbol{q}}(x)^\top\boldsymbol{\phi}_{\boldsymbol{w}}(x')
&=
\left[\sum_{t=1}^T q_t(x)\,\mathbf e_{\ell_t(x)}\right]^\top\!
\left[\sum_{s=1}^T w_s(x')\,\mathbf e_{\ell_s(x')}\right]
\\
&=
\sum_{t=1}^T \sum_{s=1}^T
q_t(x)\,w_s(x')\,
\mathbf e_{\ell_t(x)}^\top\mathbf e_{\ell_s(x')}^{\phantom{^\top}}
\\
&=
\sum_{t=1}^T
q_t(x)\,w_t(x')\,
\mathbf e_{\ell_t(x)}^\top\mathbf e_{\ell_t(x')}^{\phantom{^\top}}
\\
&=
\sum_{t=1}^T
q_t(x)\,w_t(x')\,
\mathbb I(\ell_t(x)=\ell_t(x'))
\\
&=
P_{\boldsymbol q,\boldsymbol w}(x,x'),
\end{align*}
where the third equality uses that $\mathbf e_{\ell_t(x)}^\top\mathbf e_{\ell_s(x')}^{\phantom{^\top}}=0$  since $\ell_t(x)\neq \ell_s(x')$ for $t\neq s$,   and the fourth equality uses $\mathbf e_a^\top \mathbf e_b^{\phantom{^\top}} = \mathbb I(a=b)$.
\end{proof}

\section{Implementation overview.}\label{app:implementation}

We implemented our framework in the open-source \href{https://github.com/JakeSRhodesLab/ForestGeom}{\textbf{forestgeom}} Python package, whose unified \texttt{ForestProximity} API organizes proximity computation into three stages.

First, users instantiate a \texttt{ForestProximity} object by providing a wrapped tree ensemble through the \texttt{forest} argument and selecting a proximity construction via \texttt{weight\_scheme} (e.g., \texttt{uniform}, \texttt{kerf}, \texttt{gap}, \texttt{oob}, \texttt{boosted}). A lightweight adapter layer abstracts backend-specific implementations and exposes the ensemble quantities required by proximity construction, such as leaf assignments, tree topology, OOB masks, in-bag multiplicities, and tree weights when available.

Second, fitting proceeds in two internal stages: unless a pre-trained ensemble is provided, the underlying ensemble is first trained, after which an internal cache of sparse leaf representations and auxiliary statistics is constructed; the public \texttt{fit} method composes these stages. According to the selected weighting scheme, proximities are either represented in sparse factored form\footnote{We stack sparse leaf-incidence maps row-wise to align with the standard row-vector convention in machine learning.}
$
P = QW^\top,
$
or obtained by applying a post hoc correction to a sparse leaf-collision representation. Here, \(Q\) denotes the query-side leaf map and \(W\) the reference-side leaf map. Symmetric constructions correspond to the special case \(Q=W\), while asymmetric schemes use distinct query and reference maps. For training-dependent schemes such as OOB or GAP, the API distinguishes train--train proximity computation from out-of-sample extension through dedicated methods.

Third, once the cache is available, the API exposes focused methods to retrieve fitted representations, construct query maps for new samples, and compute train--train or query--train proximity blocks. All maps are stored as SciPy sparse matrices and proximity evaluation relies on sparse matrix operations in the leaf space. This allows downstream methods to operate directly on sparse leaf-coordinate representations without materializing dense pairwise matrices, which is the primary source of the implementation's memory and computational efficiency.

\section{Compute resources.}\label{app:env}

Experiments were conducted on a compute cluster using the hardware detailed in Table~\ref{tab:hardware_specs}. The system has two Intel Ice Lake CPUs with 32 cores and 64 threads. It includes 48 MB of L3 cache and supports AVX-512 and Intel Deep Learning Boost.

\begin{table}[!ht]
\centering
\caption{Compute Cluster Hardware Specifications}
\label{tab:hardware_specs}
\small 
\begin{tabularx}{\columnwidth}{@{}l X@{}} 
\toprule
Component          & Specification \\ \midrule
Architecture                & x86\_64 (Intel Ice Lake) \\
CPU Model                   & 2$\times$ Intel Xeon Gold 6326 \\
Clock Speed                 & 2.90 GHz / 3.50 GHz \\
Cores/Threads               & 32 Cores / 64 Threads \\
System Memory (RAM)         & 512 GB DDR4 \\
Cache                       & L3: 48 MB; L2: 1.25 MB/core \\
Extensions                  & AVX-512, VNNI, Sha\_NI \\
Virtualization              & Intel VT-x with EPT \\ \bottomrule
\end{tabularx}
\end{table}

\section{Datasets.}\label{app:datasets} In this section we briefly describe the datasets used in our empirical experiments in Section~\ref{sec:results}. All data were kept as is without additional preprocessing. See Table~\ref{tab:dataset_summary} for a quantitative summary of their respective sample size, number of features and class count.

\begin{itemize}
    \item \textbf{Airlines} is a tabular binary classification dataset of commercial flights, where the goal is to predict departure delay status from scheduled flight attributes~\cite{vanschoren2014openml}.

    \item \textbf{Covertype} is a tabular multi-class classification dataset in which the goal is to predict the forest cover type of 30$\times$30 meter cells from cartographic variables such as elevation, slope, aspect, hillshade, wilderness area, and soil type~\cite{blackard_comparative_1999}.

    \item \textbf{Epsilon} is a synthetic binary classification benchmark designed for the Pascal Large Scale Learning Challenge, with high-dimensional tabular features~\cite{yuan2011improved}.  

    \item \textbf{FashionMNIST} is a dataset of Zalando's article images consisting of a training set of 60,000 examples and a test set of 10,000 examples. Each example is a 28x28 grayscale image, associated with a label from ten types of clothes~\cite{xiao_fashion-mnist_2017}.

    \item \textbf{Higgs} is a large-scale tabular binary classification dataset of simulated particle collision events, where the goal is to distinguish signal from background using 21 low-level detector measurements and seven high-level features engineered by physicists~\cite{baldi2014searching}.

    \item \textbf{PathMNIST} is a 28$\times$28$\times$3 color medical image dataset of colorectal cancer histology patches, derived from hematoxylin and eosin stained tissue images and organized into nine tissue classes for multi-class classification~\cite{yang2021medmnist, yang2023medmnist}.

    \item  \textbf{PBMC} contains single-cell gene expression profiles from peripheral blood mononuclear cells and is widely used to study cell-type structure in high-dimensional transcriptomic data~\cite{zheng2017massively}. We used the preprocessed version from DensVis, where the data are already represented in principal component space~\cite{narayan2020density}.

    \item \textbf{SignMNIST} is a grayscale 28$\times$28 image dataset of hand gestures representing American Sign Language letters for image classification~\cite{sign_language_mnist}.

    \item \textbf{SUSY} is a large-scale tabular binary classification dataset of simulated particle collision events, where the goal is to distinguish supersymmetric signal from background using low-level kinematic features measured by particle detectors~\cite{baldi2014searching}.

    \item \textbf{TissueMNIST} is a 28$\times$28 grayscale medical image dataset of segmented human kidney cortex cells, derived from 3D microscopy images and organized into 8 cell categories for multi-class classification~\cite{yang2021medmnist, yang2023medmnist}.

    \item \textbf{TV News} is a multimodal dataset of broadcast news videos, where the task is to identify commercial segments using audio-visual features extracted from video shots across diverse channels and formats~\cite{vyas_commercial_2014}.

\end{itemize}

% NSL-KDD+~\cite{tavallaee_detailed_2009}

% Zilionis~\cite{zilionis2019single}, already preprocessed from~\cite{narayan2020density},

% CElegans~\cite{packer2019lineage}, 

\begin{table}[t]
\centering
\small
\caption{Summary of the datasets used in the experiments. Columns report the dataset name, training set size, test set size (when a predefined train/test split is available), number of features, and number of classes.}
\begin{tabular}{lrrrr}
\toprule
Dataset & Training & Test & Features & Classes \\
\midrule
Airlines & $\sim$539K & -- & 8 & 2 \\
% CElegans & $\sim$86K & -- & 100 & 36 \\
Covertype & $\sim$581K & -- & 54 & 7 \\
Epsilon & 400K & 100K & 2000 & 2 \\
FashionMNIST & 60K & 10K & 784 & 10 \\
Higgs & 11M & -- & 28 & 2 \\
% NSL-KDD+ & $\sim$149K & $\sim$23K & 40 & 5 \\
PathMNIST & $\sim$97K & $\sim$7K & 2352 & 9 \\
PBMC & $\sim$69K & -- & 50 & 11 \\
SignMNIST & $\sim$35K & $\sim$7K & 784 & 24 \\
SUSY & 5M & -- & 18 & 2 \\
TissueMNIST & $\sim$213K & $\sim$47K & 784 & 8 \\
TV News & $\sim$130K & -- & 234 & 2 \\
% Zilionis & $\sim$49K & -- & 306 & 20 \\
\bottomrule
\end{tabular}

\label{tab:dataset_summary}
\end{table}

\section{Separable OOB proximities.}\label{app:separable_oob}

We saw that some proximities in the existing literature are not separable (e.g., OOB in Proposition~\ref{prop:swlc}). Since our framework shows that any proximity in the SWLC family admits a sparse factorization, it is natural to consider transforming non-separable proximities into separable ones in order to inherit near-linear scaling in the number of training samples and expose the underlying leaf-incidence maps. The challenge is that such transformations are not unique, and must preserve the underlying semantics of the original proximity. We illustrate this idea with the OOB proximity.

The standard OOB proximity $P_{\mathrm{oob}}$ (Appendix~\ref{subsec:oob}) is not compatible with our sparse matrix factorization framework, as its normalization depends jointly on $(x,x')$. However, OOB remains a strong and widely used baseline, and has been shown to perform competitively in various contexts~\cite{rhodes_geometry-_2023, chen2026randomforestinducedgraphneuralnetworks}. It is therefore desirable to construct a separable variant that retains its core behavior while enabling scalable computation.

To this end, we introduce a separable surrogate, denoted $\tilde P_{\mathrm{oob}}$, which captures the essence of the standard OOB definition while being compatible with the SWLC framework. Recall the standard OOB formulation (Appendix~\ref{subsec:oob}): the context $\theta$ includes (i) per-sample, per-tree OOB indicators $o_t(\cdot)\in\{0,1\}$, and (ii) the \emph{shared} OOB tree counts
$
S(x,x')=\sum_{t=1}^T o_t(x)\,o_t(x'),
$
which record the number of trees for which \emph{both} samples are OOB. The OOB proximity is defined as
\[
P_{\mathrm{oob}}(x,x')
=
\frac{1}{S(x,x')}\sum_{t=1}^T
o_t(x)\,o_t(x')\,\mathbb{I}\!\bigl(\ell_t(x)=\ell_t(x')\bigr).
\]
We define the following separable approximation:
\[
\tilde{P}_{\mathrm{oob}}(x,x')
=
\frac{T}{S(x)\,S(x')}
\!\sum_{t=1}^T\!
o_t(x)\,o_t(x')\,
\mathbb I\!\bigl(\ell_t(x)=\ell_t(x')\bigr),
\]
where $S(x):=\sum_{t=1}^T o_t(x).$ This approximation is obtained by replacing the pair-dependent normalization $S(x,x')$ with the separable proxy $S(x)S(x')/T$. Although the quantities $o_t(x)$, $S(x)$, and $P_{\mathrm{oob}}(x,x')$ are deterministic once a specific ensemble has been realized, the following analysis interprets them as random variables induced by the ensemble generation mechanism (e.g., bootstrap resampling), with the training set held fixed. The next result shows that, under standard RF assumptions, the shared OOB count $S(x,x')$ is well approximated by its separable counterpart $S(x)S(x')/T$.

\begin{proposition}[Asymptotic separability of OOB counts]
\label{prop:oob_shared_count}
Let \(x \neq x' \in \mathcal X\) be two distinct training samples, and assume that trees are generated independently using standard bootstrap sampling. Define
\[
p_N=\left(1-\frac{1}{N}\right)^N,
\qquad
r_N=\left(1-\frac{2}{N}\right)^N.
\]
On the event \(S(x,x')>0\), the ratio
\[
\frac{\tilde P_{\mathrm{oob}}(x,x')}{P_{\mathrm{oob}}(x,x')}
=
\frac{S(x,x')}{S(x)S(x')/T}
\]
is well defined, and satisfies
\[
\frac{\tilde P_{\mathrm{oob}}(x,x')}{P_{\mathrm{oob}}(x,x')}
\xrightarrow[T\to\infty]{P}
\frac{r_N}{p_N^2}
=
1-\mathcal O\!\left(\frac{1}{N}\right).
\]
Thus, for large \(T\), the separable approximation differs from the exact OOB normalization by an asymptotically deterministic multiplicative factor, whose bias vanishes at rate \(N^{-1}\).
\end{proposition}

\begin{proof}
For a fixed tree \(t\), the indicator \(o_t(x)\) is Bernoulli with parameter
\[
\mathbb P(o_t(x)=1)=p_N=\left(1-\frac{1}{N}\right)^N,
\]
since \(x\) is OOB exactly when it is never selected among the \(N\) bootstrap draws. Likewise, for two distinct samples \(x\neq x'\),
$
o_t(x)o_t(x')
$
is Bernoulli with parameter
\[
\mathbb P(o_t(x)=1,\ o_t(x')=1)=r_N=\left(1-\frac{2}{N}\right)^N,
\]
since both samples must be absent from all \(N\) draws. Because trees are generated independently, the law of large numbers gives
\[
\frac{S(x)}{T}\xrightarrow[]{P} p_N,
\qquad
\frac{S(x')}{T}\xrightarrow[]{P} p_N,
\qquad
\frac{S(x,x')}{T}\xrightarrow[]{P} r_N.
\]
Hence, on the event \(S(x,x')>0\),
\[
\frac{\tilde P_{\mathrm{oob}}(x,x')}{P_{\mathrm{oob}}(x,x')}
=
\frac{S(x,x')/T}{(S(x)/T)(S(x')/T)}
\xrightarrow[]{P}
\frac{r_N}{p_N^2}
\]
by the continuous mapping theorem, since \(p_N>0\).

It remains to simplify the limit:
\begin{align*}
\frac{r_N}{p_N^2}
&=
\frac{(1-2/N)^N}{(1-1/N)^{2N}}
=
\left(\frac{1-2/N}{(1-1/N)^2}\right)^N \\
&=
\left(\frac{N(N-2)}{(N-1)^2}\right)^N
=
\left(1-\frac{1}{(N-1)^2}\right)^N.
\end{align*}
Using Bernoulli's inequality,
\[
\left(1-\frac{1}{(N-1)^2}\right)^N
\ge
1-\frac{N}{(N-1)^2},
\]
while trivially
\[
\left(1-\frac{1}{(N-1)^2}\right)^N \le 1.
\]
Therefore
\[
0
\le
1-\frac{r_N}{p_N^2}
\le
\frac{N}{(N-1)^2}
=
\mathcal O\!\left(\frac{1}{N}\right),
\]
which yields
\[
\frac{r_N}{p_N^2}
=
1-\mathcal O\!\left(\frac{1}{N}\right).
\]
\end{proof}

Proposition~\ref{prop:oob_shared_count} shows that the approximation error is driven entirely by a mild normalization mismatch: the true shared OOB count is asymptotically separable, up to a bias of order $1/N$. In practice, for sufficiently large forests and training sets, this mismatch is negligible, making $\tilde P_{\mathrm{oob}}$ a faithful and scalable surrogate of the standard OOB proximity. We validate this empirically in  Section~\ref{subsec:empirical_oob_separability}.

% \begin{remark}
%     Proposition~\ref{prop:oob_shared_count} does not apply to the  case $x=x'$. Indeed, in the standard OOB definition, self-similarity is deterministic:
% $
% P_{\mathrm{oob}}(x,x)=1.
% $
% Therefore, to remain consistent with the original OOB proximity, we simply set the diagonal entries of $\tilde P_{\mathrm{oob}}$ to $1$, as done in our implementation.
% \end{remark}

\section{Extended runtime and memory ablation.}\label{app:full_ablation}

\begin{figure*}[ht] %% placed here for LaTeX float positioning
    \centering
     \includegraphics[width = \textwidth]{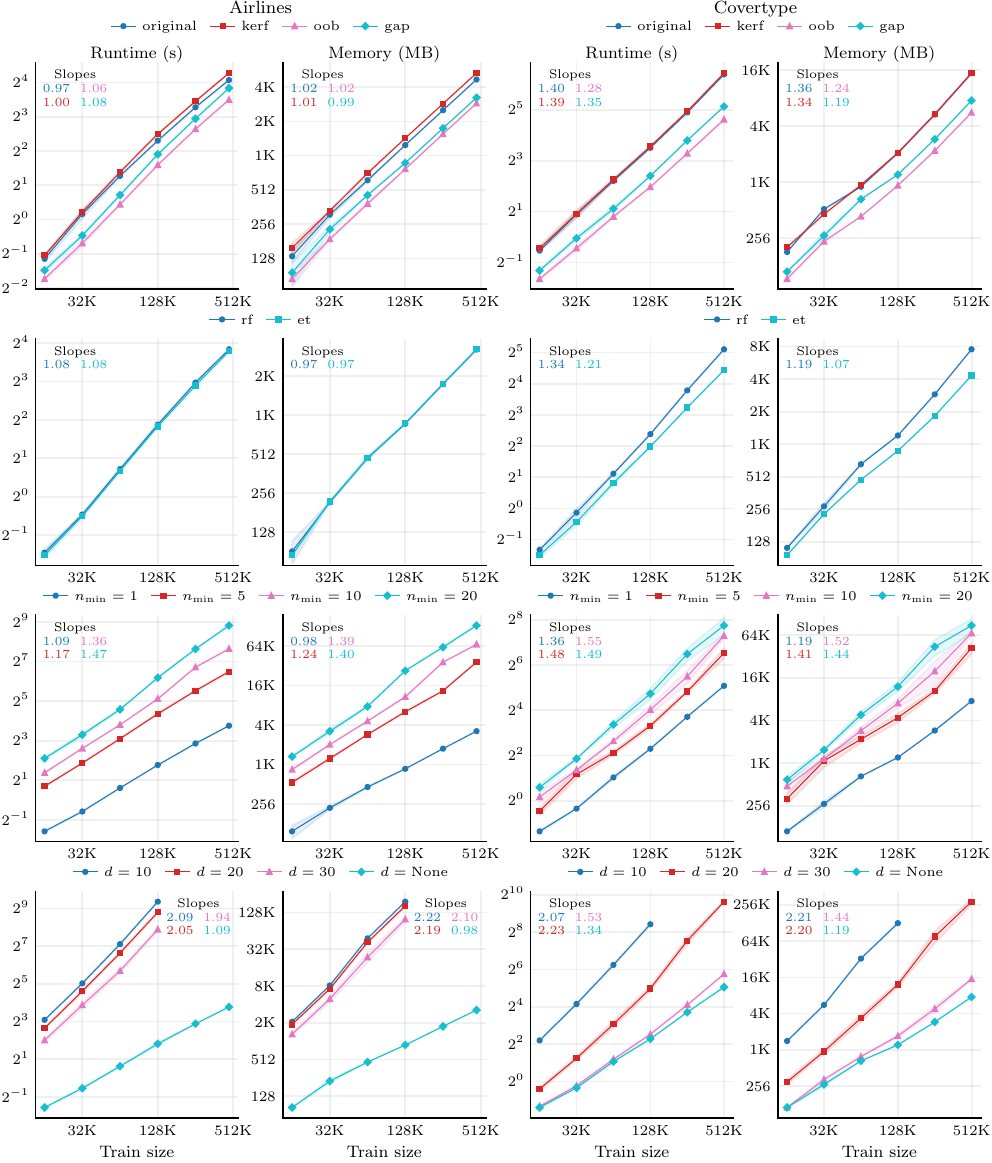}
   \caption{Log--log runtime and memory scaling of exact kernel computation with sample size under different settings: proximity method (first row), forest type (RF and ET, second row), minimum leaf size $n_{\min}$ (third row), and maximum tree depth $d$ (last row). Shaded bands indicate standard deviations, and slopes are estimated by linear regression. The left column shows ablations on the Airlines dataset, and the right column on Covertype. Near-linear scaling is observed in all cases, except under strong depth truncation, consistent with the fact that the complexity is driven by leaf interactions.}
    \label{fig:full_scaling_ablation}
\end{figure*}

In Fig.~\ref{fig:full_scaling_ablation}, we provide additional scaling ablations to complement the main results in Section~\ref{subsec:scaling_results}. In addition to the ablations across proximity methods and minimum samples per leaf on Covertype, we report the same analysis on Airlines. We also include an ablation over the forest training algorithm (RF vs.\ ET, second row), as well as a more aggressive ablation on forest hyperparameters by restricting the tree depth d (bottom row).

Overall, the main observations remain unchanged on Airlines. However, restricting the tree depth $(d \neq \text{None})$ significantly degrades scaling, both in time and memory, and can approach quadratic behavior in some cases (notably on Airlines). This is expected, as fixing the tree depth while increasing the number of samples leads to increasingly populated leaves. We therefore recommend tuning this parameter with care when scalability is a concern. In contrast, the minimum samples per leaf parameter appears less sensitive.

\section{Sanity check: accuracy of proximity-weighted classifiers.}\label{app:sanity_check_kernel_acc}

We perform a simple sanity check to verify that the kernels used in the general experiments of Section~\ref{sec:results} yield meaningful predictions. We focus on the proximity method ablation (Fig.~\ref{fig:full_scaling_ablation}) and report the corresponding kernel-weighted test accuracy for each method on Covertype and Airlines across varying training sizes. For reference, we also report the accuracy of the underlying forest.

Overall, kernel-weighted predictions closely match the forest accuracy. In particular, GAP almost perfectly recovers the forest performance. This is expected since GAP is designed to recover forest OOB predictions~\cite{rhodes_geometry-_2023}. This behavior can be beneficial or detrimental in terms of accuracy: on Covertype, GAP achieves the best performance, while on Airlines it performs worse, reflecting overfitting in the underlying forest.

In contrast, while OOB tends to be noisier~\cite{rhodes_geometry-_2023}, leading to lower accuracy on Covertype, it can improve performance on Airlines by mitigating overfitting. Our separable OOB thus provides a competitive and scalable alternative in practice.

\begin{table*}[ht]
\centering
\caption{Test accuracy of the forest predictor and kernel-weighted predictors across training sizes on Airlines and Covertype, associated with the scaling ablation experiments in Fig.~\ref{fig:full_scaling_ablation}.}
\label{tab:airlines_kernel_predict_acc}
\begin{tabular}{lccccc}
\toprule
{Train size} & {Forest} & $P_{\mathrm{gap}}$ & $\tilde P_{\mathrm{oob}}$ & $P_{\mathrm{KeRF}}$ & $P_{\mathrm{original}}$ \\
\midrule
\multicolumn{6}{c}{Airlines} \\
\midrule
16,384 & $0.629 \pm 0.002$ & $0.630 \pm 0.002$ & $0.645 \pm 0.003$ & $0.633 \pm 0.002$ & $0.645 \pm 0.002$ \\
32,768 & $0.630 \pm 0.003$ & $0.630 \pm 0.003$ & $0.649 \pm 0.003$ & $0.633 \pm 0.003$ & $0.649 \pm 0.003$ \\
65,536 & $0.629 \pm 0.003$ & $0.628 \pm 0.003$ & $0.651 \pm 0.002$ & $0.631 \pm 0.003$ & $0.650 \pm 0.003$ \\
131,072 & $0.624 \pm 0.001$ & $0.624 \pm 0.001$ & $0.652 \pm 0.002$ & $0.627 \pm 0.002$ & $0.648 \pm 0.002$ \\
262,144 & $0.618 \pm 0.002$ & $0.618 \pm 0.003$ & $0.651 \pm 0.002$ & $0.622 \pm 0.002$ & $0.646 \pm 0.003$ \\
485,444 & $0.619 \pm 0.002$ & $0.619 \pm 0.002$ & $0.645 \pm 0.002$ & $0.625 \pm 0.002$ & $0.642 \pm 0.003$ \\
\multicolumn{6}{c}{Covertype} \\
\midrule
16,384 & $0.833 \pm 0.002$ & $0.833 \pm 0.002$ & $0.791 \pm 0.001$ & $0.829 \pm 0.002$ & $0.799 \pm 0.001$ \\
32,768 & $0.863 \pm 0.002$ & $0.863 \pm 0.002$ & $0.819 \pm 0.001$ & $0.859 \pm 0.001$ & $0.826 \pm 0.001$ \\
65,536 & $0.894 \pm 0.002$ & $0.894 \pm 0.002$ & $0.850 \pm 0.003$ & $0.890 \pm 0.002$ & $0.856 \pm 0.002$ \\
131,072 & $0.920 \pm 0.002$ & $0.920 \pm 0.002$ & $0.881 \pm 0.002$ & $0.917 \pm 0.002$ & $0.885 \pm 0.002$ \\
262,144 & $0.941 \pm 0.002$ & $0.941 \pm 0.002$ & $0.907 \pm 0.001$ & $0.939 \pm 0.002$ & $0.911 \pm 0.001$ \\
522,910 & $0.957 \pm 0.001$ & $0.957 \pm 0.001$ & $0.930 \pm 0.002$ & $0.955 \pm 0.001$ & $0.932 \pm 0.002$ \\
\bottomrule
\end{tabular}
\end{table*}

\section{Additional forest leaf-based visualization.}\label{app:sign_mnist}

\begin{figure*}[ht] %% placed here for LaTeX float positioning
    \centering
     \includegraphics[width = \textwidth]{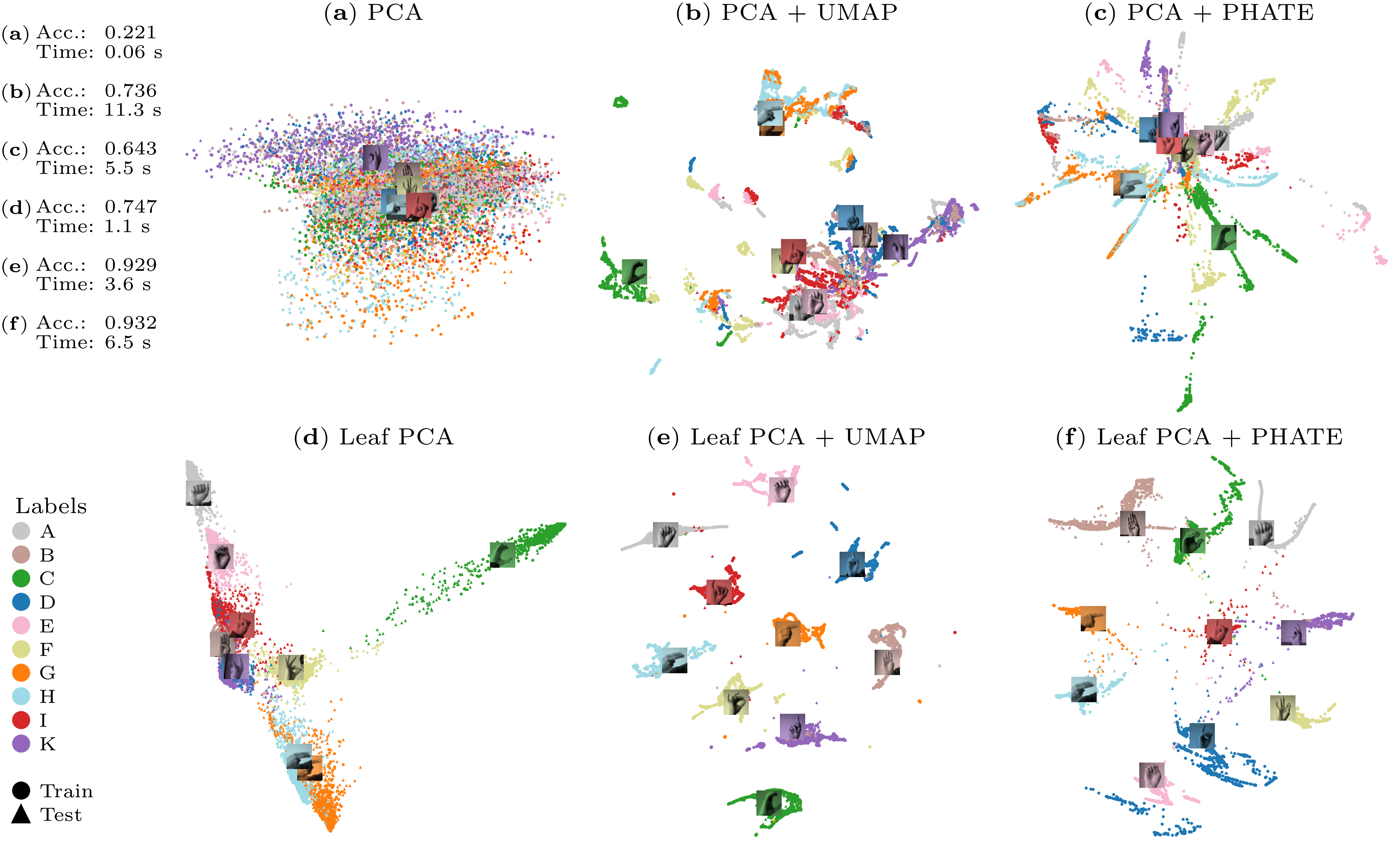}
    \caption{Two-dimensional embeddings of the SignMNIST (A--K) train and test samples using several DR pipelines applied either to the raw pixel intensity space (top row: (a) PCA, (b) PCA+UMAP, (c) PCA+PHATE) or to the leaf-incidence space (bottom row: (d)--(f)). Colors indicate class labels, circles and triangles distinguish train and test samples, and the left legend reports the corresponding runtime and test \(k\)-NN accuracy for each panel. The classwise medoid images are overlaid at their embedding locations to summarize the dominant visual patterns of each class. All methods were trained on the full predefined SignMNIST (A--K) training split. The leaf representation improves class separability across all pipelines, as reflected by higher test \(k\)-NN accuracy. Leaf PCA gives the clearest global organization, whereas raw PCA remains noisy. Raw UMAP and raw PHATE reduce some of this noise but tend to fragment classes into small disconnected islands and irregular trajectories. In contrast, Leaf UMAP and Leaf PHATE produce more coherent class-aware structure.}
    
    \label{fig:sign_mnist}
\end{figure*}

In Fig.~\ref{fig:sign_mnist}, we provide the same visual comparison as in Section~\ref{sec:manifold}, now on SignMNIST (A–K). The conclusions are similar: the leaf representation improves all DR pipelines. Leaf PCA gives the clearest global organization, with principal directions aligned with class semantics, whereas raw PCA remains highly noisy. Raw UMAP and raw PHATE reduce some of this noise, but tend to fragment classes into small disconnected islands and irregular trajectories. In contrast, Leaf UMAP and Leaf PHATE produce much more coherent structure. This improvement is also reflected in higher test $k$-NN accuracy across all leaf-based embeddings. The Leaf UMAP and Leaf PHATE embeddings remain somewhat disconnected, suggesting that larger neighborhood sizes during graph construction could further improve connectivity. More generally, our experiments indicate that Leaf PCA is a robust way to complement more complex nonlinear DR methods in leaf space by preserving a useful global view.

% \section*{Acknowledgments.}

% The acknowledgments section comes
% immediately before the references and after any appendices. It should
% be declared by $\backslash$\texttt{section*\{Acknowledgments\}} so that it is not numbered.

% The authors sincerely thank the anonymous reviewers, the Area Chairs, and the Program Chairs for their constructive feedback and their efforts in organizing the review process. We also gratefully acknowledge the assistance of the volunteers who prepared and maintained the SIAM Proceedings \LaTeX{} macros and example files.

% SIAM recommends using BibTeX
% if using BibTeX
% \clearpage
% \bibliographystyle{siamplain}
% % \bibliographystyle{unsrt}

% \bibliography{references_zotero}
\end{document}